# A Refined Analysis of *Massive Activations* in LLMs

**Louis Owen, Nilabhra Roy Chowdhury, Abhay Kumar, Fabian Güra**

BluOrion

{louis.owen, nilabhra.chowdhury, abhay.kumar, fabian.guera}@bluorion.com

March 28, 2025

## Abstract

Motivated in part by their relevance for low-precision training and quantization, massive activations in large language models (LLMs) have recently emerged as a topic of interest. However, existing analyses are limited in scope, and generalizability across architectures is unclear. This paper helps address some of these gaps by conducting an analysis of massive activations across a broad range of LLMs, including both GLU-based and non-GLU-based architectures. Our findings challenge several prior assumptions, most importantly: (1) not all massive activations are detrimental, i.e. suppressing them does not lead to an explosion of perplexity or a collapse in downstream task performance; (2) proposed mitigation strategies such as Attention KV bias are model-specific and ineffective in certain cases. We consequently investigate novel hybrid mitigation strategies; in particular pairing Target Variance Rescaling (TVR) with Attention KV bias or Dynamic Tanh (DyT) successfully balances the mitigation of massive activations with preserved downstream model performance in the scenarios we investigated. Our code is available at: `https://github.com/bluorion-com/refine_massive_activations`.

## 1 Introduction

Large Language Models (LLMs) have revolutionized natural language processing, but their internal dynamics remain poorly understood. [1] highlights a striking phenomenon: a small subset of activations exhibit values significantly larger than others, and this behavior is observed across various LLMs. These activations, referred to as "massive activations," function as implicit bias terms, concentrating attention on specific tokens. Despite these extensive contributions, several critical questions remain unanswered. In particular, there seems to be inconsistency across experimental setups–for instance, fixed bias analyses are confined to LLaMA [2], Attention KV bias mitigation is tested only on GPT-2 [3], and by default, all analyses are conducted without a leading Beginning-of-Sentence (BOS) token, with its inclusion considered only as part of ablation studies.

Massive activations are not merely an academic curiosity; they have significant practical implications for model deployment and optimization. For instance, recent investigations into Gemma-3 [4] revealed that its massive activations can cause numerical instability during inference and fine-tuning with float16 precision [5]. Specifically, the activations exceeded float16's maximum range of 65,536, resulting in infinity values and NaN gradients. This issue was traced back to the interaction between successive layer normalization operations in the decoder layers, where the output layer norm produced excessively large values.

[6] further extended the discussion of massive activations by linking them to challenges in quantization. Specifically, it highlighted how excessive magnitudes of activations—referred to as activation spikes—in Gated Liner Units (GLU) variants of Feed Forward networks (FFNs) cause severe local quantization errors, degrading the performance of quantized LLMs. The authors observed a systematic pattern: activation spikes occur in the FFNs of specific layers, particularly in early and late layers, and are dedicated to a few tokens rather than being spread across sequences. To mitigate these issues, they proposed empirical methods such as Quantization-free Module (QFeM) and Quantization-free Prefix (QFeP), and demonstrated their effectiveness in isolating activation spikes during quantization of models that exhibit massive activations pre-quantization.

| | | **Non-GLU-based Models** | | | | | | | |
|---|---|---|---|---|---|---|---|---|---|
| | **GPT-2** | | | **Phi-2-2.7B** | | | **OPT-6.7B** | | |
| Intervention | WikiText | C4 | PG-19 | WikiText | C4 | PG-19 | WikiText | C4 | PG-19 |
| Original | 30.42 | 37.61 | 53.72 | 64.40 | 87.96 | 64.16 | 10.99 | 14.65 | 13.49 |
| Set to zero | 30.57 | 37.71 | 53.85 | 64.30 | 87.93 | 64.17 | 10.99 | 14.64 | 13.48 |
| Set to mean | 30.42 | 37.61 | 53.72 | 64.40 | 87.96 | 64.16 | 10.99 | 14.65 | 13.49 |
| | **Falcon-7B** | | | **Falcon-2-11B** | | | | | |
| Intervention | WikiText | C4 | PG-19 | WikiText | C4 | PG-19 | | | |
| Original | 10.89 | 21.89 | 22.07 | 4.91 | 9.84 | 9.87 | | | |
| Set to zero | 109.65 | 154.28 | 167.36 | 5.28 | 10.84 | 11.86 | | | |
| Set to mean | 10.89 | 21.89 | 22.05 | 4.91 | 9.85 | 9.88 | | | |

| | | **GLU-based Models** | | | | | | | |
|---|---|---|---|---|---|---|---|---|---|
| | **LLaMA-2-7B** | | | **LLaMA-3.2-1B** | | | **LLaMA-3.2-3B** | | |
| Intervention | WikiText | C4 | PG-19 | WikiText | C4 | PG-19 | WikiText | C4 | PG-19 |
| Original | 5.13 | 7.65 | 8.11 | 9.07 | 14.63 | 14.93 | 7.26 | 11.88 | 11.97 |
| Set to zero | 8976.89 | 9974.18 | 6726.09 | 6249.02 | 3874.56 | 2789.20 | 18440.82 | 17798.59 | 11442.38 |
| Set to mean | 5.13 | 7.66 | 8.11 | 9.07 | 14.63 | 14.93 | 7.26 | 11.88 | 11.97 |
| | **Gemma-7B** | | | **Gemma-2-2B** | | | **Gemma-2-9B** | | |
| Intervention | WikiText | C4 | PG-19 | WikiText | C4 | PG-19 | WikiText | C4 | PG-19 |
| Original | 6.40 | 10.71 | 11.39 | 8.05 | 13.18 | 13.67 | 6.38 | 11.33 | 10.55 |
| Set to zero | $2.76 \times 10^{24}$ | $4.77 \times 10^{25}$ | $6.41 \times 10^{27}$ | 101.73 | 144.34 | 144.99 | 7.09 | 12.84 | 11.31 |
| Set to mean | 6.40 | 10.71 | 11.38 | 8.05 | 13.18 | 13.69 | 6.38 | 11.33 | 10.55 |
| | **Gemma-3-1B** | | | **Gemma-3-4B** | | | **Gemma-3-12B** | | |
| Intervention | WikiText | C4 | PG-19 | WikiText | C4 | PG-19 | WikiText | C4 | PG-19 |
| Original | 9.79 | 15.48 | 16.56 | 6.88 | 11.74 | 11.69 | 5.49 | 10.03 | 8.99 |
| Set to zero | 10.21 | 16.07 | 21.01 | 172.25 | 301.86 | 325.20 | 5.52 | 10.09 | 9.16 |
| Set to mean | 9.79 | 15.48 | 6.90 | 11.79 | 11.81 | 15.92 | 5.49 | 10.05 | 9.03 |
| | **OLMo-7B-0724** | | | **OLMo-2-1124-7B** | | | **Phi-4-14B** | | |
| Intervention | WikiText | C4 | PG-19 | WikiText | C4 | PG-19 | WikiText | C4 | PG-19 |
| Original | 7.14 | 10.60 | 10.29 | 5.76 | 12.35 | 10.29 | 6.08 | 11.88 | 9.63 |
| Set to zero | 15549.53 | 13510.87 | 7573.01 | - | - | - | 6.23 | 12.13 | 10.04 |
| Set to mean | 7.14 | 10.60 | 10.29 | - | - | - | 6.08 | 11.88 | 9.63 |
| | **Mistral-7B-v0.3** | | | | | | | | |
| Intervention | WikiText | C4 | PG-19 | | | | | | |
| Original | 4.99 | 8.48 | 8.47 | | | | | | |
| Set to zero | $2.62 \times 10^{7}$ | $3.74 \times 10^{7}$ | $2.92 \times 10^{7}$ | | | | | | |
| Set to mean | 4.99 | 8.48 | 8.47 | | | | | | |

Table 1: **Perplexity scores for various pre-trained LLMs subject to massive activation intervention *with the BOS token included.*** A "-" indicates that no massive activations were found. Results are color-coded to indicate the level of impact: purple denotes highly detrimental effects, while orange signifies medium detrimental effects. For consistency, we used a context length of 4096 tokens for all models, except for GPT-2 and OPT-6.7B, where context lengths of 1024 and 2048 tokens were used, respectively, due to model limitations. By default, all analyses were conducted using float16 precision, except for Gemma-7B and all Gemma-3 models, which required float32. For Gemma-7B, this was due to the original weights being in float32, while for the Gemma-3 models, the excessively high activation magnitudes exceeded the range supported by float16.

This paper addresses some gaps in literature by conducting a comprehensive analysis of massive activations across a broad range of LLMs, including both GLU-based and non-GLU-based architectures. We investigate several mitigation strategies aimed at managing massive activations, such as Attention KV bias [1], Target Variance Rescaling (TVR) [7], and Dynamic Tanh (DyT) [8]. These strategies are evaluated not only for their effectiveness in reducing activation magnitudes, but also for their impact on downstream task performance and attention dynamics. By systematically analyzing these approaches, we aim to advance the understanding of massive activations and their implications for LLM design. Furthermore, this work identifies unresolved questions and proposes directions for future research, paving the way for more robust and architecture-agnostic strategies to manage activation dynamics in LLMs.

## 2   Prior Work

The authors of [1] identified a phenomenon where a small subset of activations exhibits significantly larger values than others, which they termed *massive activations*. Specifically, they defined an LLM as having massive activations if the maximum magnitude of its hidden states exceeds 100 and is at least 1,000 times greater than the median magnitude of the hidden states. Mathematically, this can be expressed as:

$$\max(|\mathbf{h}|) > 100 \quad \text{and} \quad \max(|\mathbf{h}|) \geq 1000 * \text{median}(|\mathbf{h}|), \tag{1}$$

where $\mathbf{h}$ represents the hidden states of the model, and $|\cdot|$ denotes the element-wise absolute value operation.

Additionally, they observed the following three key characteristics of massive activations:

**1. Constant Values Across Layers**  Massive activations remain largely constant throughout most intermediate layers. They emerge in the initial layers and diminish in the final layers. For some models (e.g., LLaMA2-7B and LLaMA2-13B [9]), they emerge rapidly within a single layer, while for others (e.g., Phi-2 [10] and OPT [11]), they accumulate gradually over many layers.

**2. Fixed Location**  In terms of the sequence dimension, massive activations are located on specific types of tokens. Tokens associated with massive activations are typically starting word tokens, delimiter tokens, or tokens with weak semantics (e.g., "and," "of," "from"). These tokens are few in number, but play a critical role in shaping model behavior. In terms of the feature dimension, massive activations are consistently present in very few fixed dimensions.

**3. Fixed Bias**  Massive activations act as important bias terms, not just as redundant activations with no effect; the model seems to re-purposing the tokens linked to them to store these biases. This phenomenon is observed through *intervention analysis*: setting massive activation values to zero severely degrades model performance, while replacing them with their mean values (computed over $N$ input samples) does not cause significant harm. This indicates that their values are constant and input-agnostic.

Massive activations also seem to be closely connected to self-attention mechanisms. A stark contrast in attention patterns emerges when comparing layers before and after the appearance of massive activations. Specifically, in layers following the first occurrence of massive activations, attention becomes concentrated on tokens associated with these activations. For a multi-token input sequence which triggers massive activations, the tokens associated with massive activations exhibit drastically different key and value states compared to the key and value states of other tokens in the same sequence. The high activation values skew the denominator term of normalization layers such as LayerNorm [12] and RMSNorm [13]. This results in drastically different feature representations for the tokens associated with massive activation. These tokens are utilized to form a constant bias term during attention computation, which acts as an *implicit bias* term in the model. This conclusion was drawn from a study on *attention output decomposition*, where the attention output is decomposed into two parts: value updates from the tokens $\mathcal{C}$ where attention is concentrated, and value updates aggregated from other tokens:

$$\text{Attention}(Q, K, V)_k = \sum_{i \leq k} p_i^k v_i = \sum_{i \in \mathcal{C}} p_i^k v_i + \sum_{i \notin \mathcal{C}} p_i^k v_i, \tag{2}$$

where $p_i^k$ is the attention distribution of query token $k$ to token $i$, and $v_i$ is the value state of token $i$. The analysis showed that value updates from $\mathcal{C}$ are nearly identical across tokens, effectively functioning as additive bias terms despite not being explicitly imposed.

The authors furthermore proposed a method to eliminate massive activations by augmenting the self-attention mechanism with additional key and value embeddings, which are explicitly designed to act as biases. Rather than re-purposing existing tokens in the input sequence, this method introduces additional *learnable* parameters $\mathbf{k}', \mathbf{v}' \in \mathbb{R}^d$ for each attention head. Specifically, given the input query, key, and value matrices $Q, K, V \in \mathbb{R}^{T \times d}$, where $T$ denotes the sequence length and $d$ represents the embedding dimension, the augmented attention mechanism with explicit biases is computed as follows:

$$\text{Attention}(Q, K, V; \mathbf{k}', \mathbf{v}') = \text{softmax}\left(\frac{Q\left[K^T \quad \mathbf{k}'\right]}{\sqrt{d}}\right)\begin{bmatrix} V \\ \mathbf{v}'^T \end{bmatrix}, \tag{3}$$

where $\mathbf{k}'$ and $\mathbf{v}'$ are concatenated with the key and value matrices $K$ and $V$, respectively. This formulation allows the proposed attention mechanism to serve as a drop-in replacement for standard attention, without requiring modifications to other components of the Transformer architecture. Notably, this approach effectively adds an extra "token" on-the-fly, ensuring that the biases are incorporated seamlessly into the computation. However, this extra "token" is transient in nature: it only persists until the matrix multiplication between $QK$ and $V$ is completed. Subsequently, due to the properties of matrix multiplication, this "token" is no longer explicitly retained in the output, as its influence is implicitly integrated into the resulting attention values.

## 3    Experiments

Our goal in this paper was to systematically assess the existence, characteristics, and impact of massive activations while addressing several gaps in prior analyses. We adopted the definition proposed by [1], where an LLM is considered to have massive activations if the maximum magnitude value of the hidden states exceeds 100 and is at least 1000 times greater than the median magnitude value of the hidden states (see also Equation 1). This definition ensures consistency with prior studies while enabling reproducibility across diverse architectures. Using this criterion, we ran experiments in two key areas: **intervention analysis** and **mitigation strategies**.

It's important to note that by default prior work [1] conducted analyses without including the BOS token. However, this approach may not align with how most models are trained and supposed to be configured for inference, as inputs typically include the BOS token. While they observed that massive activations persisted even when the BOS token was included, their findings were limited to specific models (LLaMA2-7B, LLaMA2-13B, and Mixtral-8x7B [17]). They acknowledged this limitation and suggested extending their analysis to a broader range of models as part of future work. In this study, we evaluated each model under two conditions: with and without the BOS token. This setup allowed us to explore how architectural differences and input variations influenced the emergence and behavior of massive activations.

### 3.1    Intervention analysis

One key claim by [1] that we sought to validate is that massive activations *always* act as fixed "bias" terms, essentially re-purposing specific tokens to store these biases. If suppressed, these biases can negatively impact model performance, particularly in terms of perplexity and downstream task performance. In short, this hypothesizes that massive activations are *always detrimental*. It is worth noting that the original authors based their conclusions primarily on experiments conducted with LLaMA2-7B and LLaMA2-13B. However, there is no guarantee that the same behavior generalizes to other architectures.

We adopted the intervention methodology introduced in [1]. Specifically, we modified the inference process of LLMs by intervening on massive activations at one layer. For hidden states exhibiting massive activations, we manually set these activations to chosen fixed values (e.g., zero or their mean). The altered hidden state was then fed into the next layer, allowing computation to proceed as normal. Following the original study, we evaluated perplexity on WikiText [18], C4 [19], and PG-19 [20].

- **Set to zero.** To assess the importance of massive activations, we removed them by setting their values to zero in the hidden state at the point where they first appeared. This effectively eliminated massive activations. If these activations are indeed detrimental, we would expect the measured perplexity to increase significantly. Conversely, if they are not detrimental, the perplexity should remain relatively stable.

- **Set to mean.** To test whether small variances in massive activation values contribute to their role, we replaced these values with their empirical means. If the perplexity remains similar after this intervention, it suggests that the values of massive activations are constant and input-agnostic, functioning similarly to bias terms. Unlike the original approach, which focused solely on token indices from a particular sentence (e.g., "Summer is Warm. Winter is Cold.")[1], we first identified all locations of massive activations across the model from 100 samples of RedPajama [25]. These activations were then categorized into two groups: starting tokens (e.g., index 0) and non-starting tokens (index > 0). This categorization allowed us to systematically analyze how massive activations behave at different token positions, effectively controlling for the significant differences in scale of starting and non-starting positions. We then repeated this procedure on the test datasets WikiText, PG-19, and C4, organizing once again massive activations into the two buckets of starting and non-starting tokens, and replace the corresponding tokens in a sample with the respective mean values derived from RedPajama[2].

The models under investigation include both GLU-based and non-GLU-based architectures. Among the GLU-based models, we examine LLaMA-2-7B [9], LLaMA-3.2-1B and LLaMA-3.2-3B [30], Mistral-7B-v0.3 [31], Gemma-7B [14], Gemma-2-2B and Gemma-2-9B [32], OLMo-7B-0724 [15], OLMo-2-1124-7B [33], and Phi-4 [34]. For non-GLU-based models, we include GPT-2 [3], Falcon-7B and Falcon-2-11B [16], OPT-6.7B [11], and Phi-2 [10].

---

[1]See code here.
[2]See code here.

## 3.2 Mitigation Strategies

**Methods**

Another claim we investigated is that augmenting self-attention with explicit key and value biases (see Equation (3)) can eliminate massive activations. However, it is important to note that this conclusion by [1] is based solely on experiments conducted on the GPT-2 architecture. The generalizability of this mitigation strategy to other architectures seems to be untested so far.

In this work, we extended the evaluation of mitigation strategies to a larger and more mainstream architecture, namely LLaMA-1B. We tested various strategies, such as Attention KV Bias, which repurposes existing tokens to act as biases; Target Variance Rescaling (TVR), which rescales weights to control extreme values; Dynamic Tanh (DyT), which is a replacement for all layer norm modules; and hybrid approaches that combine TVR with other strategies, such as KV Bias + TVR and DyT + TVR. The choice of methods to evaluate is motivated by the intuition (and some prior indication) that correlate massive activations with large weight variance (TVR) and layer normalization (DyT), respectively.

**Benchmarks**

Unlike prior studies that relied primarily on validation loss as the sole evaluation metric, we adopted a comprehensive evaluation framework encompassing a diverse set of tasks designed to measure various aspects of language understanding and reasoning. These tasks include commonsense reasoning benchmarks such as HellaSwag [26], PIQA [22], SIQA [27], and WinoGrande [23]; question-answering tasks such as TriviaQA [28]; and multi-step reasoning tasks such as ARC-Challenge (ARC-C) and ARC-Easy (ARC-E) [24].

**Training Setup**

**GPT-2**. We used the default GPT-2-124M model configuration and the default GPT-2 tokenizer with 1024 context length as available on Huggingface[3]. We utilized only the FineWebEdu-Deduplicated subset of the SmolLM [29] dataset, training the model for a total of 50 billion tokens. We always prepend BOS token to every document. All weights were initialized from a normal distribution of zero mean and 0.02 standard deviation, with special residual initialization strategy proposed in [3] applied. All training with the GPT-2 architecture was performed using the hyperparameters presented in Appendix 6.1.

**LLaMA-1B**. As for the one-billion-parameter LLaMA model, the training data consists of the SmolLM dataset, which combines three diverse sources: FineWebEdu-Deduplicated, Cosmopedia-V2 and Python-Edu; we randomly sample 100 billion tokens (100BT) from this corpus. The context length is fixed at 2048 tokens per sequence. We always prepend BOS token in every document. All 2D modules weights, except stated otherwise, were initialized from a normal distribution of zero mean and 0.006 standard deviation, with Layer Index Rescaling (LIR) applied [7]. Details for the hyperparameters, model configuration, and tokenizer are presented in Appendix 6.2.

## 4 Results

### 4.1 Not all Massive Activations are detrimental

First, we observe that not all massive activations are detrimental, especially for most of the non-GLU LLMs (see Table 1 and Table 6 for results with and without BOS token, respectively). While some models experienced significantly worse perplexity when massive activations were suppressed by setting their values to zero, others showed minimal or no degradation in performance. Despite this variability, we have yet to identify a clear indicator that distinguishes between detrimental and non-detrimental massive activations.

However, one *weak* pattern emerges: models which have quickly increasing massive activations across the first few layers, quickly decreasing massive activations for the final few layers, and almost constant massive activations for the middle layers, tend to exhibit more detrimental effects. An example for this is LLaMA-2-7B as illustrated in Figure 1a. On the other hand, Falcon-7B displays extreme emergence and diminishment, but with slightly gradual steps; its massive activations are only moderately detrimental (see Figure 1b). On the other end of the spectrum, Falcon-2-11B exhibits multiple step-wise increases on intermediate layers; its massive activations appear to be non-detrimental (see Figure 1c). However, there are exceptions to to the

---

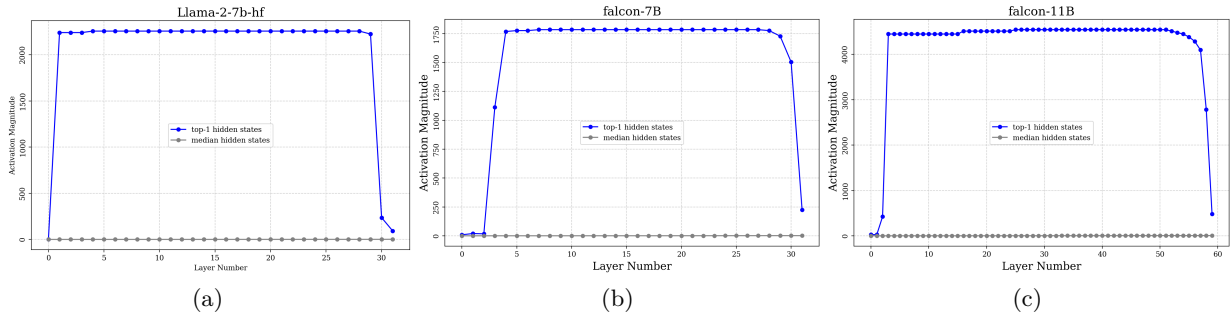

Figure 1: **Top activation magnitude across layers for (a) Llama-2-7B; (b) Falcon-7B; (c) Falcon-2-11B.** The input sentence is "Summer is warm. Winter is cold" with BOS token included.

pattern: For example, Gemma-7B with BOS token has detrimental massive activation (see Table 1), but the pattern of the top activation magnitude across layers does not follow the standard pattern demonstrating a spike in magnitude on the last layer (see Figure 10d). Top activation magnitude plots for all investigated LLMs with and without BOS token can be found in Appendix 6.4 and 6.5.

## 4.2 BOS Token is significant for some LLMs

Second, we find that the BOS token plays a significant role in Gemma, though its influence varies across different model generations and sizes. Figure 2 illustrates this effect: when the BOS token is excluded, massive activations are not observed in Gemma-7B, Gemma-2-2B, and Gemma-2-9B. However, when the BOS token is included, massive activations emerge in these models. Additionally, while massive activations persist in the Gemma-3 models (Gemma-3-1B, Gemma-3-4B, and Gemma-3-12B) even without the BOS token (see Figures 8g, 8h, 8i), their magnitudes are significantly reduced compared to when the BOS token is present (see Figures 10g, 10h, 10i). Notably, from a perplexity perspective (see Tables 1 and 6), excluding the BOS token also leads to worsened performance across all Gemma models, underscoring the importance of the BOS token in this model family.

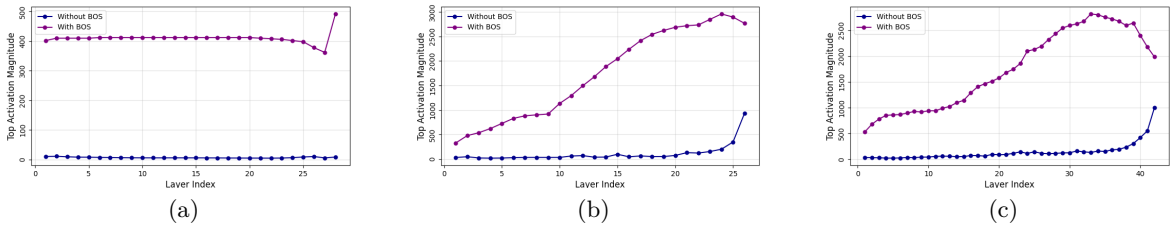

Figure 2: **Comparison of top activation magnitude across layers for (a) Gemma-7B; (b) Gemma-2-2B; (c) Gemma-2-9B; with and without BOS token**. The input sentence is "Summer is warm. Winter is cold".

## 4.3 Attention KV Bias is not a general Mitigation Strategy

Our investigation into the effectiveness of attention key-value bias as a mitigation strategy yielded mixed results. We retrained multiple GPT-2 and LLaMA-1B models from scratch following the setup describe in section 3.2. For GPT-2, we were able to reproduce the findings from prior studies, confirming that KV bias effectively reduces the magnitude of massive activations. Figure 3a illustrates the significant reduction in top activation magnitudes when KV bias is applied. However, downstream task performance on benchmarks such as HellaSwag consistently underperformed throughout the whole training process (see Figure 3b). This underscores the limitations of relying solely on validation loss as a performance metric.

In contrast, the same KV bias formulation failed to mitigate massive activations in LLaMA-1B. As shown in Figure 3c, the top activation magnitudes remained largely unchanged; downstream task performance underperformed as well (see Table 7 and Figure 3d). This suggests that KV bias as a mitigation strategy for massive activations is not universally effective across architectures.

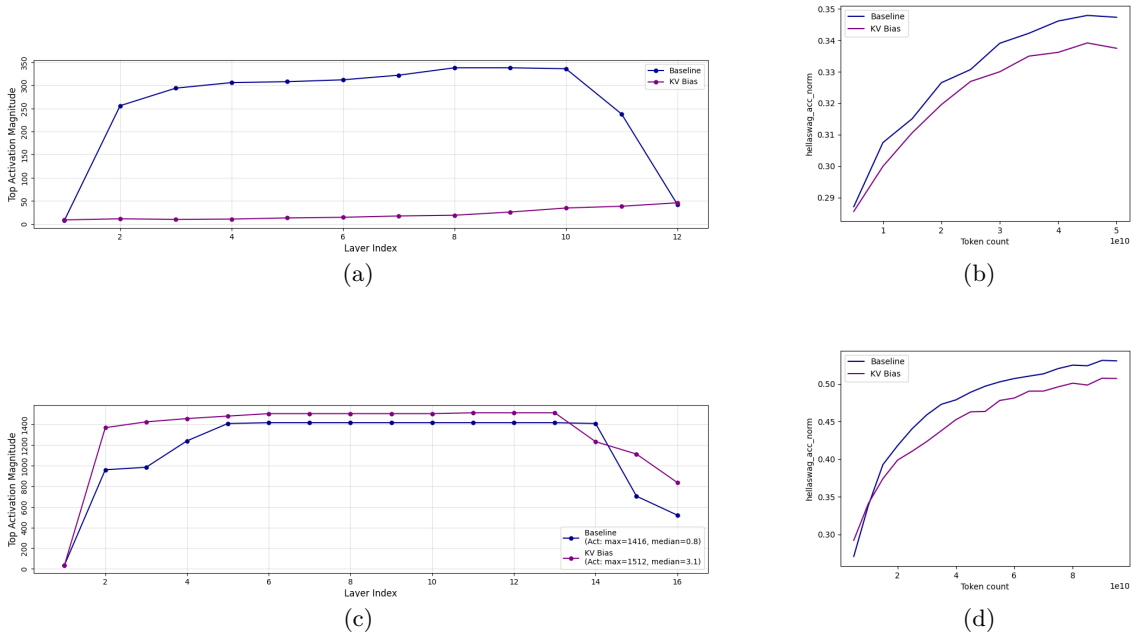

Figure 3: **Comparison of top activation magnitude across layers for (a) GPT-2 and (c) LLaMA-1B with and without KV Bias. Evolution of HellaSwag scores for (b) GPT-2 and (d) LLaMA-1B with and without KV Bias.** The input sentence is "Summer is warm. Winter is cold" with BOS token included. We presented the corresponding analysis without BOS token in Appendix 6.7.3.

To better understand the impact of KV bias on attention patterns, we analyzed the distribution of attention weights before and after applying the mitigation strategy. Figure 13 shows 2D heatmaps of attention weights for GPT-2 with the BOS token included, illustrating the average attention logits across all heads, tokens, and layers. The results reveal that attention becomes concentrated on the BOS token starting from layer 3 (note that massive activations first appear in layer 2; see Figure 3a). While KV bias successfully reduces the magnitude of massive activations, it does not eliminate attention concentration. Instead, the extra "token" continues to attract a significant proportion of attention across layers (see Figure 14). A similar pattern is observed when the BOS token is excluded (see Figures 11 and 12).

In contrast, the effect of KV bias on LLaMA-1B differs from its impact on GPT-2, particularly when the BOS token is included. For LLaMA-1B, attention is initially concentrated on the extra "token" in the layers immediately following the emergence of massive activations (specifically, from layer 3 to layer 6). Beyond layer 6, attention shifts back to the BOS token (see Figure 22). This highlights the key role played by the BOS token in LLaMA-1B. When the BOS token is included, the baseline model exhibits strong attention concentration on the BOS token across all layers (see Figure 21). However, this pattern changes when the BOS token is excluded. As shown in Figure 15, the attention distribution becomes less concentrated and splits between two tokens: the starting token and the full-stop token.

## 4.4 Alternative Mitigation Strategies

We explored alternative mitigation strategies trying to address both massive activations and attention concentration while preserving downstream task performance. We focused our experiments on LLaMA-1B, using it as a representative model to evaluate the effectiveness of various approaches. While this scope allows for an in-depth analysis of LLaMA-1B's behavior, future work should extend these investigations to other architectures to validate the generalizability of our findings.

|  | Mitigates Massive Activations? | Max Activation Magnitude | Median Activation Magnitude | Mean Downstream Task Accuracy |
|---|---|---|---|---|
| Baseline | × | 1416 | 0.8 | 50.3 |
| KV Bias | × | 1512 | 3.1 | 49.6 |
| DyT | ✓ | 47 | 1.8 | 49.9 |
| TVR | ∼ | 235 | **0.2** | **52.5** |
| KV Bias + TVR | ✓ | 158 | 0.7 | 52.0 |
| DyT + TVR | ✓ | **45** | 1.3 | 50.3 |

Table 2: **Summary of the effectiveness of various mitigation strategies for massive activations in LLaMA-1B, along with their impact on downstream task performance.** Breakdown of downstream tasks performance is presented in Appendix 6.8.

### 4.4.1 Dynamic Tanh (DyT)

We studied the effects of replacing the RMSNorm layers in the model with Dynamic Tanh (DyT) [8]. To ensure convergence, we incorporated embedding scaling as proposed by the authors. Following their recommendations, we set the hyperparameters based on the width and depth of LLaMA-1B. Specifically, for the DyT layers in the attention modules, we used $\alpha = 1.0$, while for the DyT layer before the final projection layer, we used $\alpha = 0.5$. The embedding scaler was initialized with $\sqrt{d}$, where $d$ denotes the embedding dimension, as suggested in the paper. We initialized all 2D module weights in the decoder layer with zero mean and a standard deviation of 0.02, with LIR [7] applied. This specific value was determined through hyperparameter tuning experiments, where we observed that other initialization standard deviations resulted in drastically worse downstream task performance.

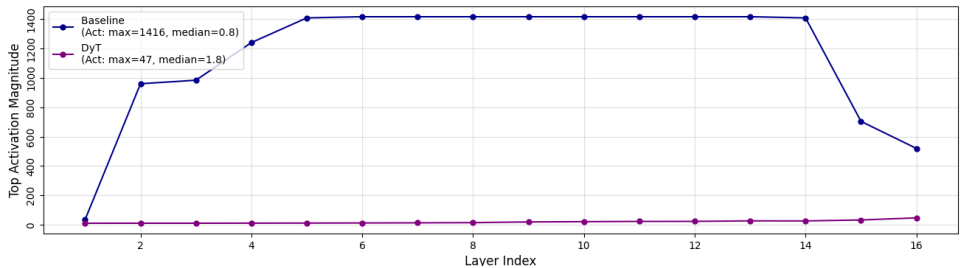

Figure 4: **Comparison of top activation magnitude across layers for LLaMA-1B baseline and with DyT.** The input sentence is "Summer is warm. Winter is cold" with BOS token included. We presented the corresponding analysis without BOS token in Appendix 6.7.3.

Figure 4 illustrates the reduction in top activation magnitudes, confirming DyT's effectiveness in mitigating massive activations. However, this improvement came at a cost: downstream task performance deteriorated, though it remained slightly better than with KV bias alone (see Table 2). Furthermore, attention concentration persisted even with DyT, albeit at a reduced level (see Figures 19 and 25). This finding challenges the hypothesis that RMSNorm or LayerNorm play a key role in enabling tokens associated with massive activations to exhibit drastically different key and value states [1]. These differences were previously suggested to lead to the formation of a constant bias term during attention computation. By entirely replacing normalization operations, DyT was expected to drastically disrupt this mechanism. However, the empirical evidence suggests otherwise, indicating that the relationship between normalization, massive activations, and attention concentration is more complex than previously hypothesized.

### 4.4.2 Target Variance Rescaling (TVR)

Target Variance Rescaling (TVR), introduced in [7], aides in stabilizing the variance of model weights during pre-training. In their experiments on LLaMA-1B, TVR successfully reduced extreme activation values by approximately tenfold. However, lowering activation magnitudes does not necessarily equate to removing massive activations as per our definition. To validate TVR's effectiveness, we conducted experiments using a target standard deviation of 0.01, as recommended in the original paper, and applied it to LLaMA-1B.

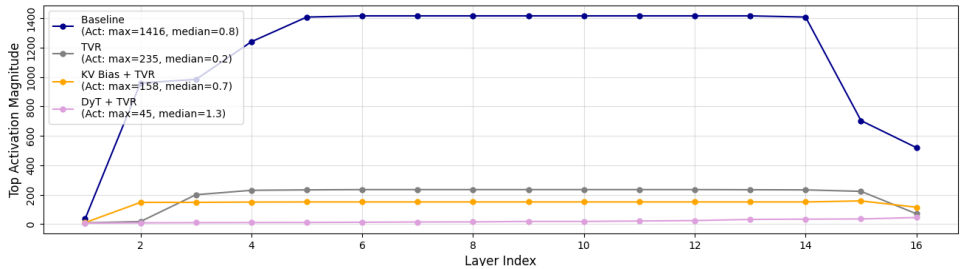

Figure 5: **Comparison of top activation magnitude across layers for LLaMA-1B with TVR, KV Bias + TVR, and DyT + TVR**. The input sentence is "Summer is warm. Winter is cold" with BOS token included. We presented the corresponding analysis without BOS token in Appendix 6.7.3.

The results, shown in Figure 5, confirm that TVR significantly reduces extreme activation values. However, massive activations persist under our definition (the maximum activation magnitude value is still 1000 times greater than the median magnitude value), indicating that TVR alone does not fully eliminate this phenomenon. Despite this limitation, downstream task performance remained comparable to or even slightly improved over the baseline (see Table 2). These findings suggest that TVR can mitigate massive activations to some degree without compromising model performance. Nevertheless, attention concentration persists, as illustrated in Figures 17 and 23.

To further enhance TVR's effectiveness, we combined it with KV bias to create a hybrid mitigation strategy. As shown in Figure 5, this combination achieved an even greater reduction in extreme activation values compared to using either method alone, hence resulting in the elimination of massive activation as per our definition. Downstream tasks performance also remained stable, with some tasks even showing improvements over the baseline (see Table 2). These results demonstrate that combining TVR and KV bias can effectively mitigate some undesirable effects of massive activations without compromising model performance. Nevertheless, attention concentration still persisted with similar pattern found in incorporating KV bias without TVR (Figures 18 and 24).

Finally, we combined DyT with TVR to explore whether TVR could recover the lost downstream task performance observed with DyT alone. In this hybrid approach, we initialized all 2D module weights in the decoder layer with zero mean and a standard deviation of 0.02, while applying LIR as before. Additionally, we set the TVR target standard deviation to 0.02.

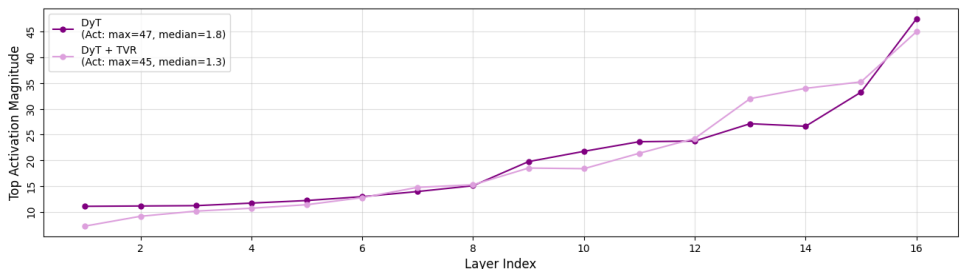

Figure 6: **Comparison of top activation magnitude across layers for LLaMA-1B between DyT and DyT + TVR**. The input sentence is "Summer is warm. Winter is cold" with BOS token included. We presented the corresponding analysis without BOS token in Appendix 6.7.3.

This combination proved highly effective: Table 2 demonstrates that downstream task performance recovers to levels comparable to the baseline. More importantly, DyT + TVR continues to suppress extreme activation values effectively (see Figures 5 and 6). Additionally, while attention concentration is further reduced compared to only incorporating DyT without TVR, it still persists (Figures 20 and 26). Consistent with the hypothesis proposed by the original authors [1], these findings suggest that LLMs have a strong intrinsic propensity to form attention concentration patterns during pre-training.

In a nutshell, these findings highlight the potential of hybrid strategies to achieve a balanced mitigation of massive activations. While individual methods like TVR, KV bias, and DyT offer partial solutions, their combinations—such as KV bias + TVR, and DyT + TVR—provide more comprehensive approaches that preserve model performance while addressing undesirable massive activations.

### 4.5 Other Characteristics

Our analysis of massive activations reveals both consistencies and contradictions with prior findings across various LLM architectures. One key characteristic previously identified is that massive activation values tend to be mostly constant across layers [1]. However, we observed deviations from this pattern in certain models. For instance, models such as Gemma-2-2B, Gemma-3-1B, Gemma-3-4B, and Gemma-3-14B display a "uptrend" increase in activation magnitudes across layers (see Figures 8 and 10). This behavior might be influenced by architectural design choices, as these models employ both pre- and post-layernorm operations in their MLP and attention modules. In contrast, other models that rely solely on pre-layernorm do not exhibit this trend, suggesting that normalization strategies could play a role in shaping activation dynamics.

Another characteristic pertains to the fixed sequence dimension locations of massive activations. Prior study [1] suggested that these activations are primarily associated with starting, delimiter, or weak semantic tokens. However, our findings reveal exceptions in models like Gemma-3-4B, Gemma-3-12B, and Falcon-7B. For these models, massive activations are linked not only to starting or delimiter tokens, but also to other tokens within the sequence, such as "polished", "mass", "cold". This challenges the assumption that massive activations are strictly tied to specific token types, highlighting the need for architecture-specific investigations.

Despite these contradictions, several characteristics remain consistent with prior observations. Massive activations are consistently small in number, and occur in very few fixed embedding positions across different models. This alignment reinforces the robustness of earlier findings while underscoring the importance of understanding how these activations interact with model architecture and training dynamics. For additional insights and relevant illustrations, we refer readers to the supplementary materials available in our repository: `https://github.com/bluorion-com/refine_massive_activations`.

## 5    Conclusion and Future Work

In this work, we conducted a systematic analysis of massive activations in LLMs, addressing gaps in prior studies by evaluating their existence, characteristics, and impact across a diverse range of architectures. Our findings challenge several previously published assumptions about massive activations, and highlight the need for architecture-specific mitigation strategies.

First, we demonstrated that not all massive activations are detrimental to model performance. While some models exhibit significant degradation in perplexity and downstream task performance when massive activations are suppressed, others remain almost completely unaffected. This suggests that the role of massive activations is more nuanced than previously thought, warranting further investigation into distinguishing factors between detrimental and non-detrimental cases.

Second, our results indicate that previously presented mitigation strategies are not universally generalizable. In particular, Attention KV bias fails to mitigate massive activations effectively in LLaMA-1B, underscoring its limitations in certain architectures. However, on the LLaMA-1B model specifically, when combined with Target Variance Rescaling (TVR), this approach successfully suppresses extreme activation values without compromising downstream task performance. This highlights the potential of hybrid strategies in addressing massive activations while preserving model performance.

Furthermore, we explored the DyT (Dynamic Tanh) architecture as an alternative approach to mitigating massive activations. While DyT effectively removes these activations, it comes at the cost of reduced performance on downstream tasks in our experiments. Interestingly, combining DyT with TVR recovers downstream task performance, suggesting that TVR plays a crucial role in balancing activation suppression and functional preservation.

These findings underscore the importance of developing architecture-aware mitigation strategies and exploring combinations of techniques to achieve optimal results. Moving forward, several directions for future work emerge:

- Investigating the underlying mechanisms that differentiate detrimental from non-detrimental massive activations, potentially through fine-grained analysis of their distribution and persistence across layers.

- Extending the evaluation of hybrid strategies such as KV Bias + TVR and DyT + TVR to a broader range of architectures and tasks to validate their generalizability.

- Analyzing in more depth how the attention concentration phenomenon exhibits under multiple different mitigation techniques on a broader range of LLMs.

Future research in this area has the potential to unlock new insights into model behavior and pave the way for more reliable and scalable AI systems. Notably, given the relevance of massive activations to low-precision training and quantization, our findings provide valuable directions for addressing numerical stability challenges in these contexts.

# 6  Appendix

## 6.1  Training Hyperparameters for GPT-2

The hyperparameters details for GPT-2 are presented in Table 3.

| Hyperparameter | Value |
|---|---|
| Optimizer | Fused AdamW ($\beta_1 = 0.9$, $\beta_2 = 0.95$, $\epsilon = 1 \times 10^{-7}$) |
| Learning Rate Schedule | Linear warmup followed by cosine decay |
| Max Learning Rate | $6 \times 10^{-4}$ |
| End Learning Rate | $6 \times 10^{-5}$ |
| Warmup Tokens | 1 billion tokens (1BT) |
| Weight Decay | 0.1 (AdamW implementation) |
| Global Batch Size | 2048 |
| Precision | Mixed Precision BFloat16 |
| Gradient Clipping | 1.0 |
| Dropout rate | 0.0 |
| Bias | not used |

Table 3: **GPT-2 Training Hyperparameters**

## 6.2  Training Hyperparameters, Model Config, and Tokenizer Details for LLaMA-1B

Table 4 shows details of the model architecture used in our experiment setup. The tokenizer used in this work is a truncated version of the Llama3 tokenizer, which includes a vocabulary size of 65,536 tokens. Training hyperparameters details is presented in Table 5.

| Parameter | Value |
|---|---|
| Hidden Size | 2048 |
| Intermediate Size | 5440 |
| Number of Hidden Layers | 16 |
| Number of Attention Heads | 16 |
| Number of Key-Value Heads | 16 |
| Activation Function | SwiGLU |
| RMSNorm Epsilon | $1 \times 10^{-5}$ |
| Vocabulary Size | 65,536 |
| Maximum Position Embeddings | 2048 |

Table 4: **LLaMA-1B Model Configuration**

| Hyperparameter | Value |
|---|---|
| Optimizer | Fused AdamW ($\beta_1 = 0.9$, $\beta_2 = 0.95$, $\epsilon = 1 \times 10^{-7}$) |
| Learning Rate Schedule | Linear warmup followed by cosine decay |
| Max Learning Rate | $6 \times 10^{-4}$ |
| End Learning Rate | $6 \times 10^{-5}$ |
| Warmup Tokens | 1 billion tokens (1BT) |
| Weight Decay | 0.1 (AdamW implementation) |
| Global Batch Size | 4032 |
| Precision | Mixed Precision BFloat16 |

Table 5: **LLaMA-1B Training Hyperparameters**

### 6.3 Intervention Analysis Results without BOS token

Table 6 shows the intervention analysis results without BOS token for various LLMs.

| Non-GLU-based Models | | | | | | | | |
|---|---|---|---|---|---|---|---|---|
| **GPT-2** | | | **Phi-2-2.7B** | | | **OPT-6.7B** | | |
| Intervention | WikiText | C4 | PG-19 | WikiText | C4 | PG-19 | WikiText | C4 | PG-19 |
| Original | 30.22 | 37.33 | 53.28 | 62.13 | 83.70 | 62.85 | 10.97 | 14.54 | 13.35 |
| Set to zero | 30.21 | 37.31 | 53.29 | 62.26 | 83.96 | 62.94 | 10.98 | 14.54 | 13.35 |
| Set to mean | 30.21 | 37.31 | 53.28 | 61.94 | 83.73 | 62.88 | 10.97 | 14.54 | 13.35 |
| **Falcon-7B** | | | **Falcon-2-11B** | | | | | |
| Intervention | WikiText | C4 | PG-19 | WikiText | C4 | PG-19 | | | |
| Original | 10.88 | 21.91 | 21.98 | 4.87 | 9.78 | 9.79 | | | |
| Set to zero | 117.27 | 163.36 | 174.86 | 5.29 | 10.90 | 11.98 | | | |
| Set to mean | 10.89 | 21.92 | 22.13 | 4.87 | 9.79 | 9.85 | | | |
| GLU-based Models | | | | | | | | |
| **LLaMA-2-7B** | | | **LLaMA-3.2-1B** | | | **LLaMA-3.2-3B** | | |
| Intervention | WikiText | C4 | PG-19 | WikiText | C4 | PG-19 | WikiText | C4 | PG-19 |
| Original | 5.12 | 7.62 | 8.08 | 9.07 | 14.46 | 14.88 | 7.28 | 11.83 | 11.97 |
| Set to zero | 8115.29 | 8625.69 | 5745.95 | 7364.78 | 5669.25 | 3865.99 | 22144.77 | 22513.59 | 14938.57 |
| Set to mean | 5.12 | 7.63 | 8.08 | 9.09 | 14.78 | 15.02 | 7.29 | 11.85 | 11.99 |
| **Gemma-7B** | | | **Gemma-2-2B** | | | **Gemma-2-9B** | | |
| Intervention | WikiText | C4 | PG-19 | WikiText | C4 | PG-19 | WikiText | C4 | PG-19 |
| Original | $3.11 \times 10^8$ | $9.61 \times 10^7$ | $6.00 \times 10^7$ | 57.14 | 65.50 | 48.23 | 122.60 | 235.10 | 254.67 |
| Set to zero | - | - | - | - | - | - | - | - | - |
| Set to mean | - | - | - | - | - | - | - | - | - |
| **Gemma-3-1B** | | | **Gemma-3-4B** | | | **Gemma-3-12B** | | |
| Intervention | WikiText | C4 | PG-19 | WikiText | C4 | PG-19 | WikiText | C4 | PG-19 |
| Original | 11.67 | 18.26 | 20.99 | 8.47 | 14.31 | 15.33 | 14.35 | 24.91 | 25.66 |
| Set to zero | 11.72 | 18.50 | 21.05 | 266.79 | 540.92 | 460.32 | 15.87 | 27.92 | 27.51 |
| Set to mean | 11.65 | 18.26 | 20.99 | 9.00 | 15.25 | 15.99 | 14.91 | 26.75 | 26.31 |
| **OLMo-7B-0724** | | | **OLMo-2-1124-7B** | | | **Phi-4-14B** | | |
| Intervention | WikiText | C4 | PG-19 | WikiText | C4 | PG-19 | WikiText | C4 | PG-19 |
| Original | 7.10 | 10.55 | 10.24 | 5.73 | 12.29 | 10.24 | 6.06 | 11.87 | 9.57 |
| Set to zero | 9014.45 | 9500.15 | 5497.85 | - | - | - | 6.18 | 12.13 | 10.00 |
| Set to mean | 7.10 | 10.55 | 10.25 | - | - | - | 6.05 | 11.86 | 9.58 |
| **Mistral-7B-v0.3** | | | | | | | | |
| Intervention | WikiText | C4 | PG-19 | | | | | | |
| Original | 5.00 | 8.44 | 8.43 | | | | | | |
| Set to zero | $1.80 \times 10^6$ | $2.06 \times 10^6$ | $1.59 \times 10^6$ | | | | | | |
| Set to mean | 4.98 | 8.47 | 8.45 | | | | | | |

Table 6: **Perplexity scores for various pre-trained LLMs subject to massive activation intervention *with the BOS token excluded.*** A "-" indicates that no massive activations were found. Results are color-coded to indicate the level of impact: purple denotes highly detrimental effects, while orange signifies medium detrimental effects. For consistency, we used a context length of 4096 tokens for all models, except for GPT-2 and OPT-6.7B, where context lengths of 1024 and 2048 tokens were used, respectively, due to model limitation. By default, all analyses were conducted using float16 precision, except for Gemma-7B and all Gemma-3 models, which required float32. For Gemma-7B, this was due to the original weights being in float32, while for the Gemma-3 models, the excessively high activation magnitudes exceeded the range supported by float16.

### 6.4 Top Activation Magnitude Plots without BOS token

Figure 7 and 8 illustrate the top activation magnitude plots for non-GLU and GLU-based LLMs without BOS token, respectively.

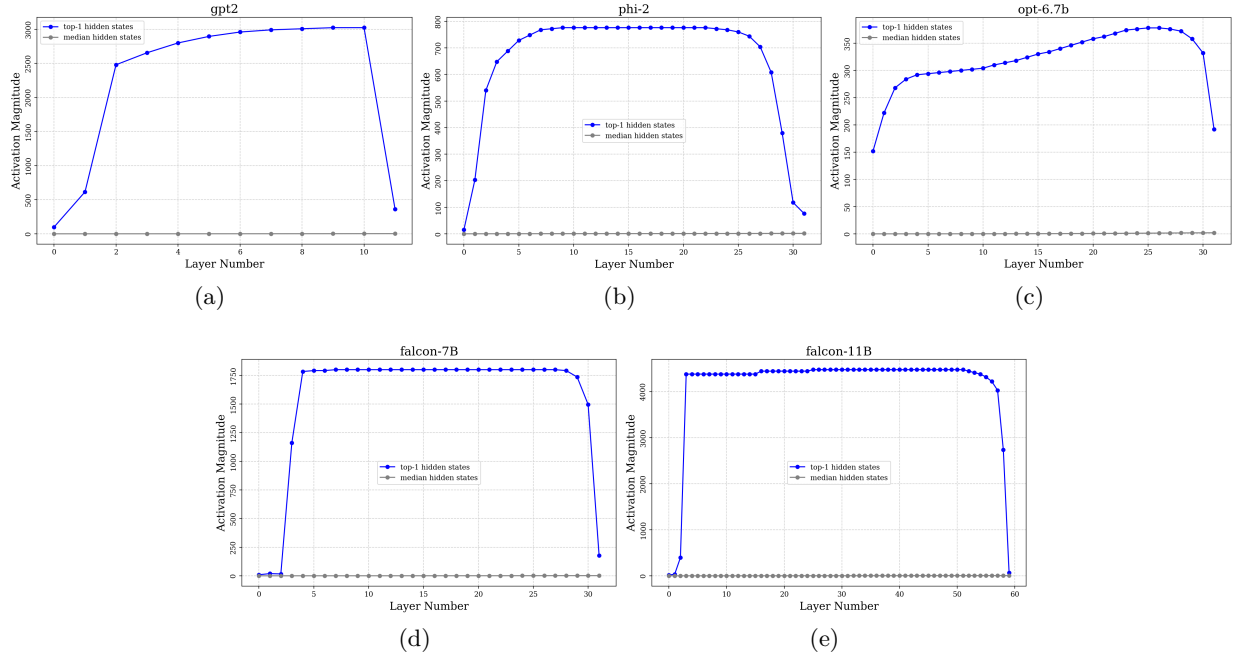

Figure 7: **Top activation magnitude across layers for Non-GLU-based LLMs: (a) GPT-2; (b) Phi-2; (c) OPT-6.7B; (d) Falcon-7B; (e) Falcon-2-11B.** The input sentence is "Summer is warm. Winter is cold" with BOS token excluded.

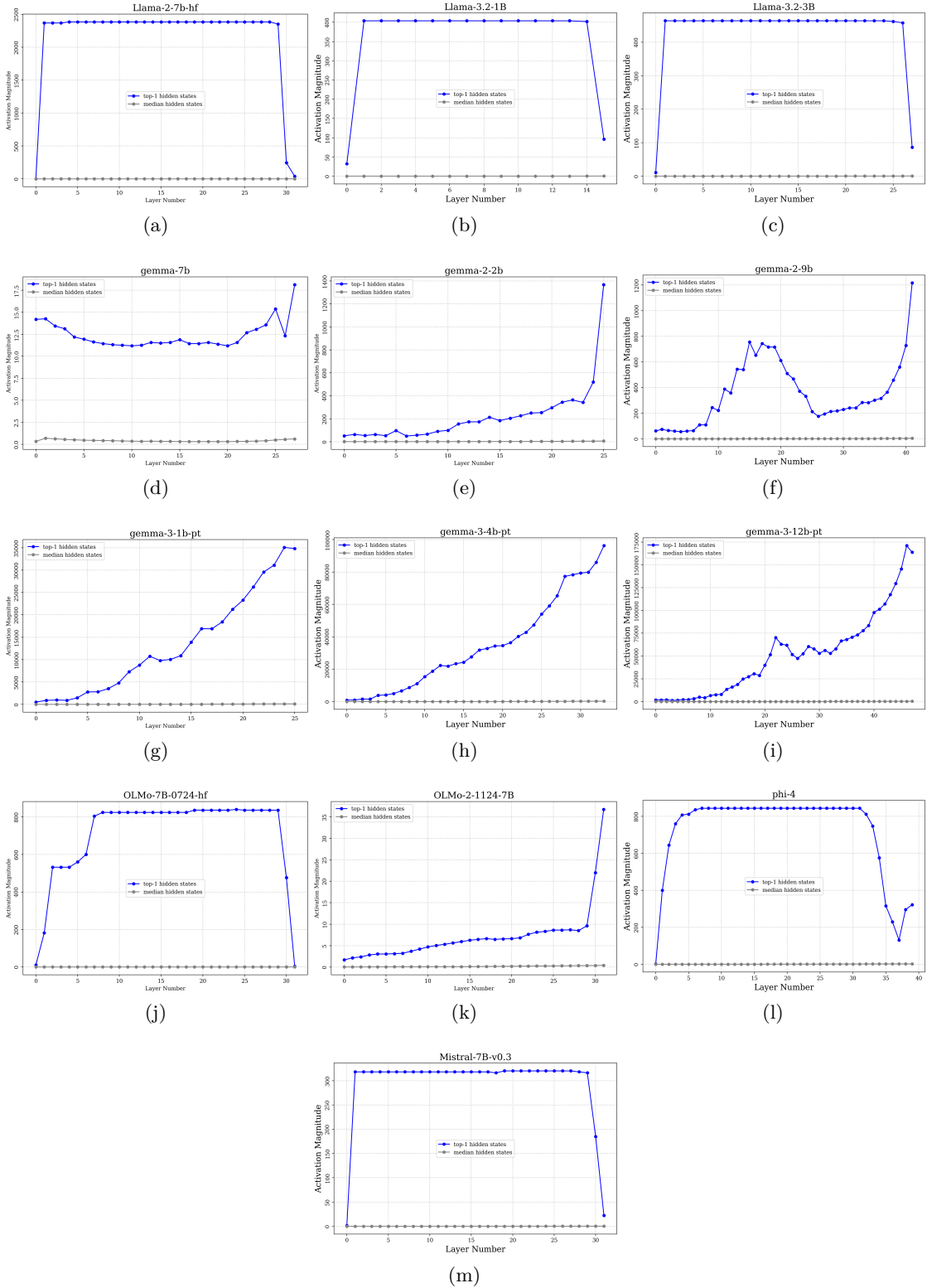

Figure 8: **Top activation magnitude across layers for GLU-based LLMs: (a) LLaMA-2-7B; (b) LLaMA-3.2-1B; (c) LLaMA-3.2-3B; (d) Gemma-7B; (e) Gemma-2-2B; (f) Gemma-2-9B; (g) Gemma-3-1b; (h) Gemma-3-4b; (i) Gemma-3-12b; (j) OLMo-7B-0724; (k) OLMo-2-1124-7B; (l) Phi-4; (m) Mistral-7B-v0.3.** The input sentence is "Summer is warm. Winter is cold" with BOS token excluded.

### 6.5 Top Activation Magnitude Plots with BOS token

Figure 9 and 10 illustrate the top activation magnitude plots for non-GLU and GLU-based LLMs with BOS token, respectively.

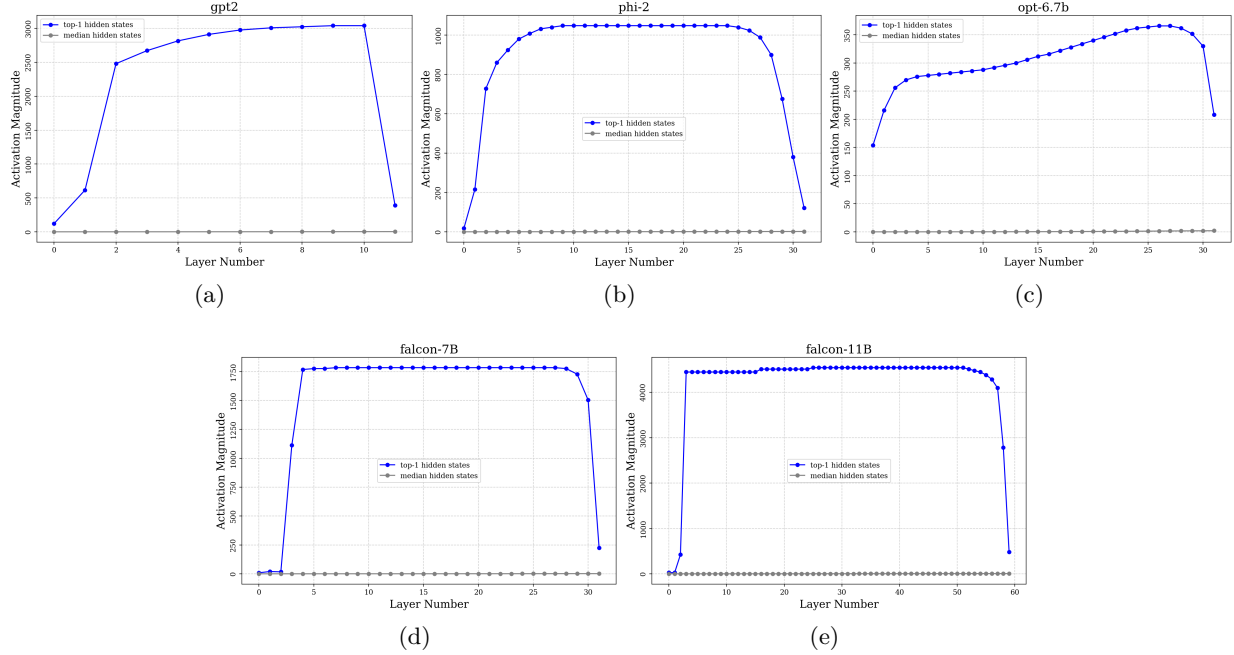

Figure 9: **Top activation magnitude across layers for Non-GLU-based LLMs: (a) GPT-2; (b) Phi-2; (c) OPT-6.7B; (d) Falcon-7B; (e) Falcon-2-11B**. The input sentence is "Summer is warm. Winter is cold" with BOS token included.

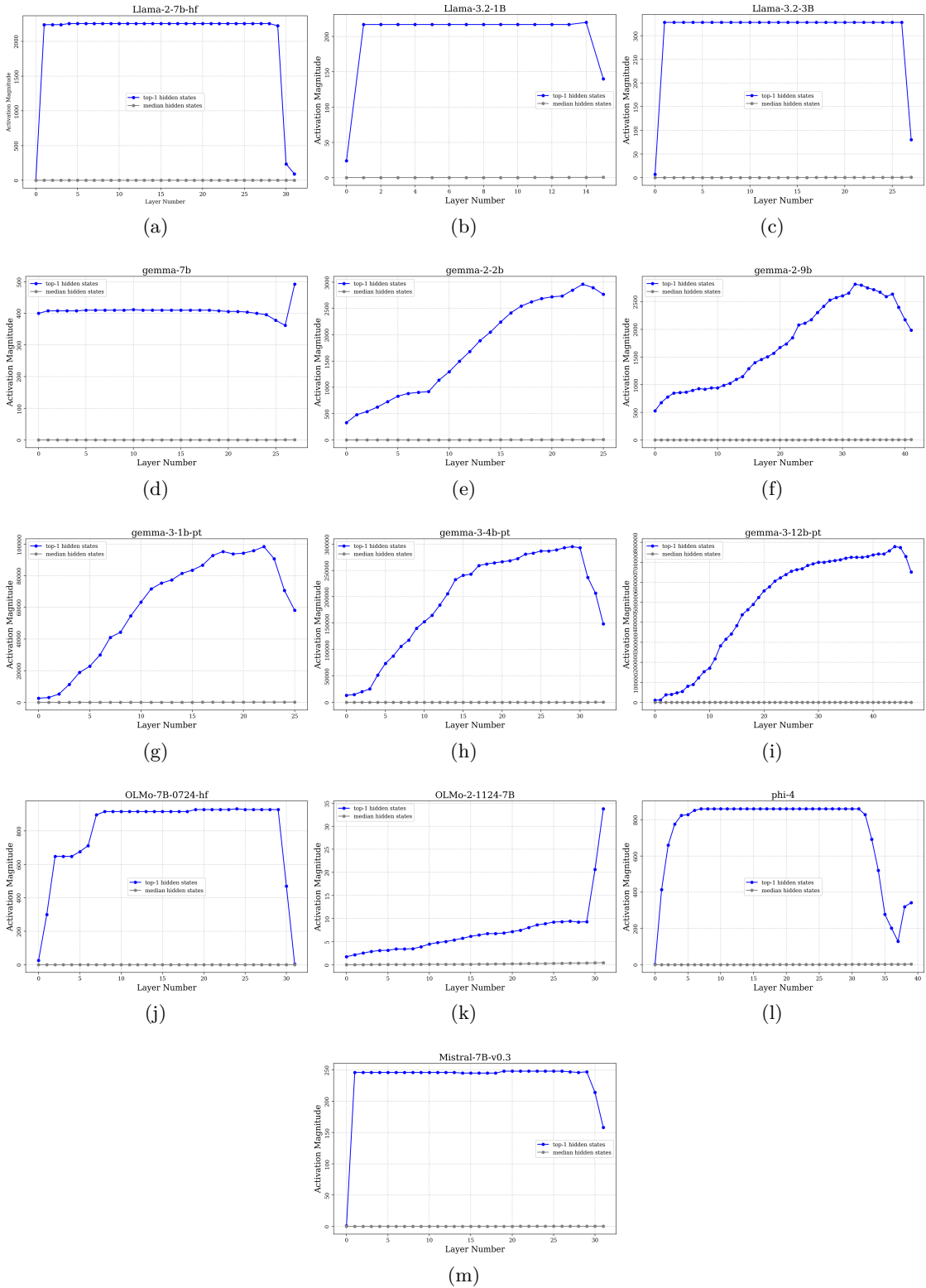

Figure 10: **Top activation magnitude across layers for GLU-based LLMs: (a) LLaMA-2-7B; (b) LLaMA-3.2-1B; (c) LLaMA-3.2-3B; (d) Gemma-7B; (e) Gemma-2-2B; (f) Gemma-2-9B; (g) Gemma-3-1b; (h) Gemma-3-4b; (i) Gemma-3-12b; (j) OLMo-7B-0724; (k) OLMo-2-1124-7B; (l) Phi-4; (m) Mistral-7B-v0.3.** The input sentence is "Summer is warm. Winter is cold" with BOS token included.

## 6.6 Self Attention Plots for Retrained GPT-2

### 6.6.1 Self Attention Plots for Retrained GPT-2 without BOS token

Figures 11 - 12 illustrates the self attention distribution for the GPT-2 model that we trained from scratch. BOS token is *not* included in all of these visualizations. Note that in GPT-2 the BOS token shares the same token id with the EOS token.

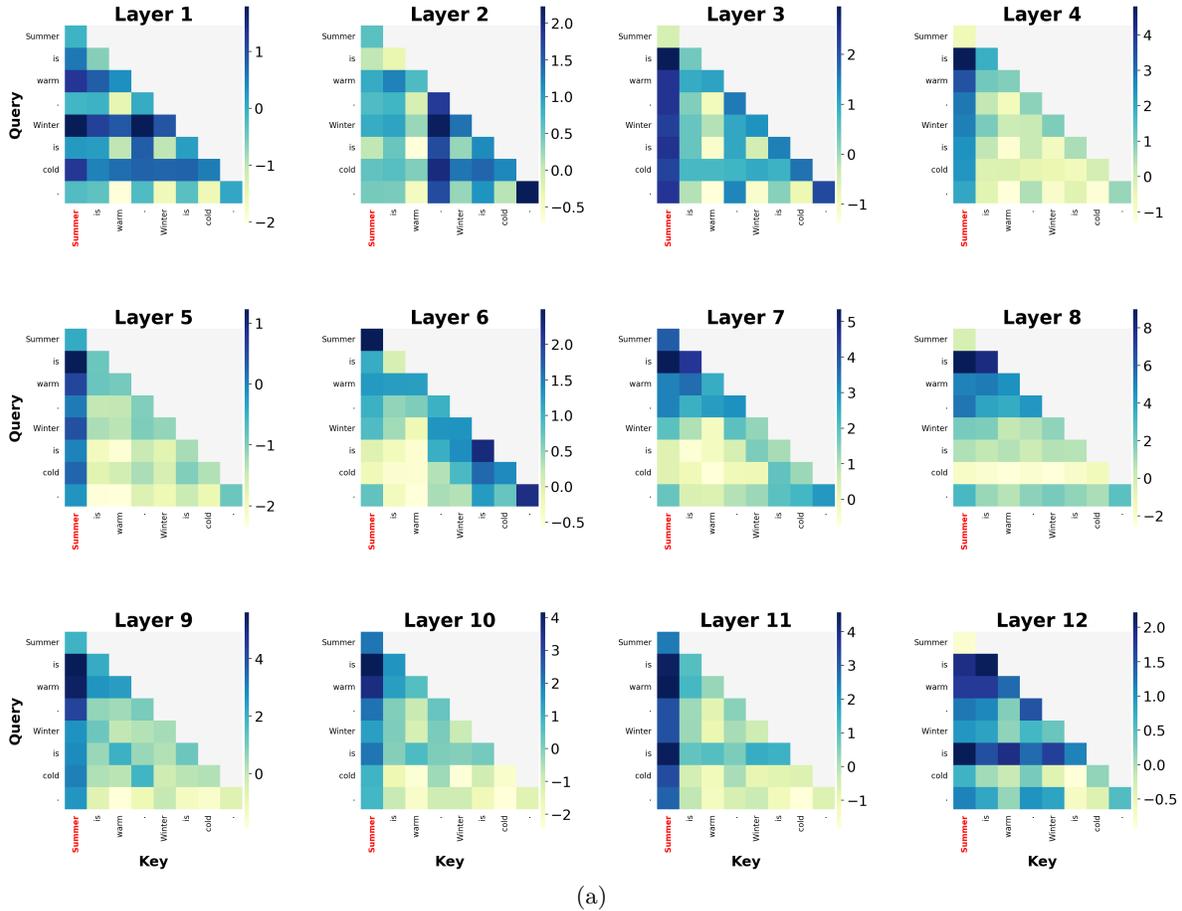

(a)

Figure 11: **Average attention logits over all heads for GPT-2 (baseline) without BOS token.**

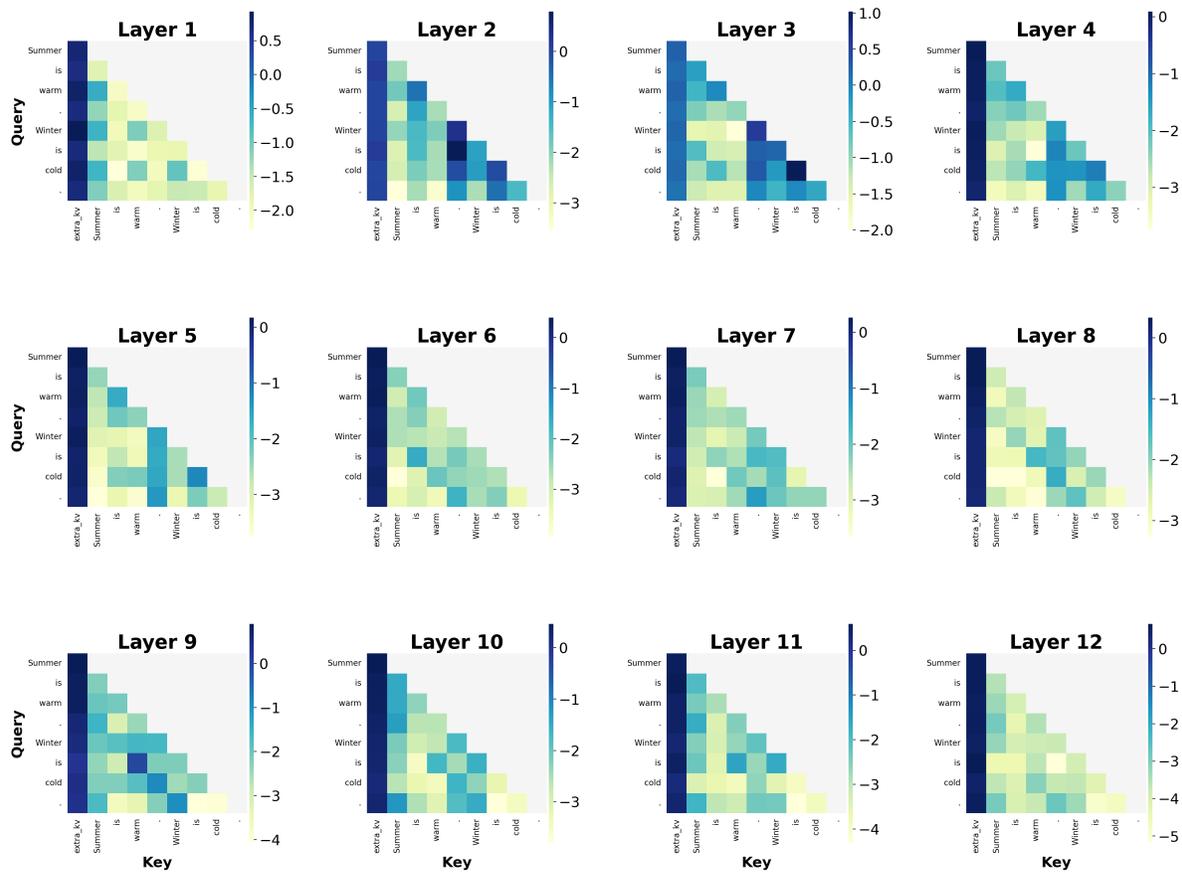

(a)

Figure 12: **Average attention logits over all heads for GPT-2 (with KV Bias) without BOS token.**

### 6.6.2 Self Attention Plots for Retrained GPT-2 with BOS token

Figures 13 - 14 illustrates the self attention distribution for the GPT-2 model that we trained from scratch. BOS token is included in all of these visualizations. Note that in GPT-2 the BOS token shares the same token id with the EOS token.

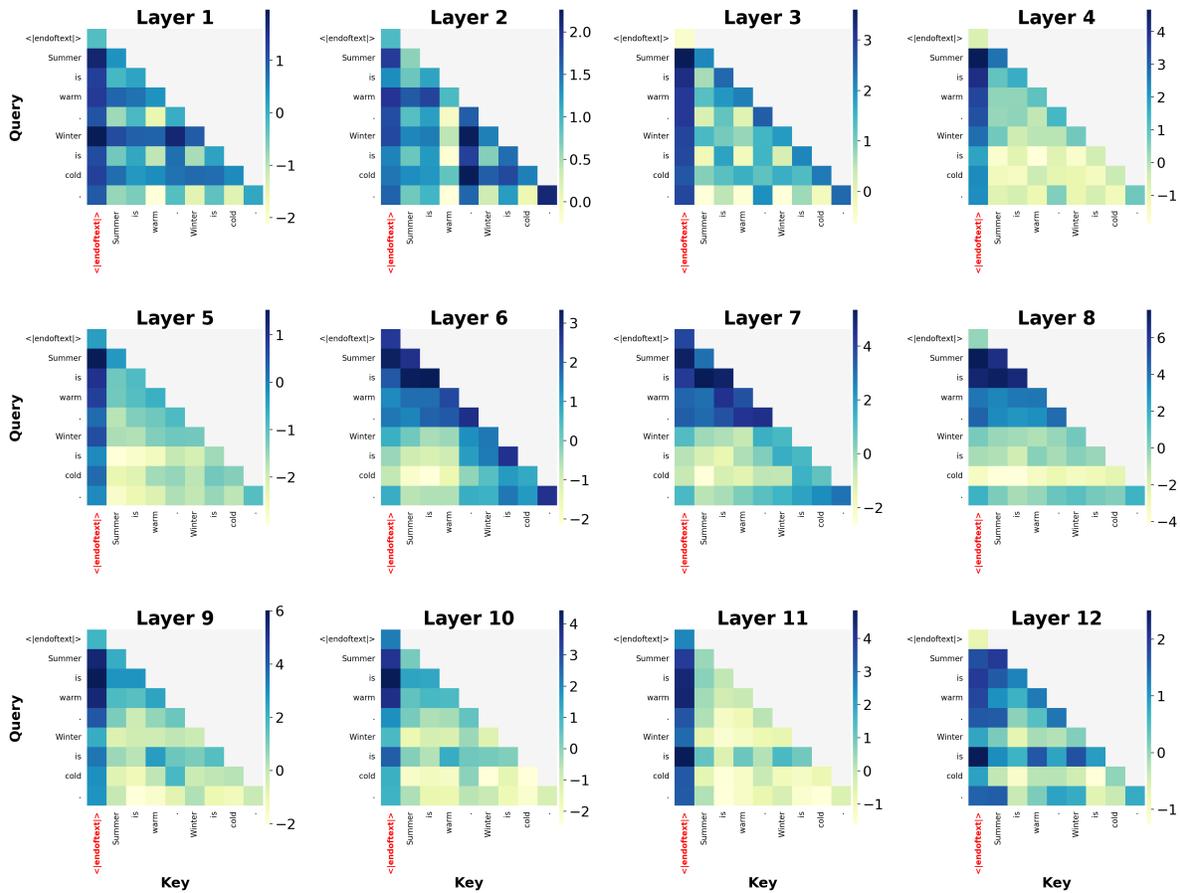

Figure 13: **Average attention logits over all heads for GPT-2 (baseline) with BOS token.**

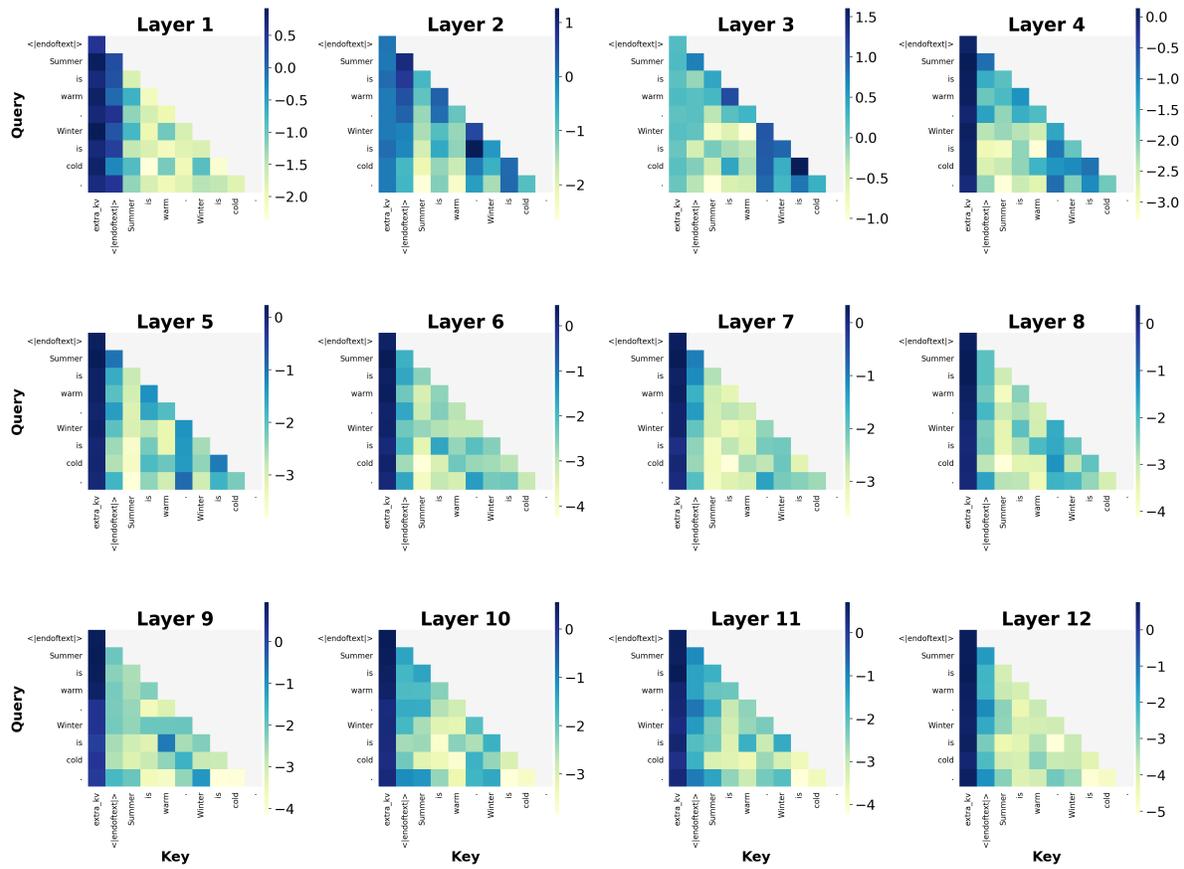

Figure 14: **Average attention logits over all heads for GPT-2 (with KV Bias) with BOS token.**

## 6.7 Self Attention Plots for Retrained LLaMA-1B

### 6.7.1 Self Attention Plots for Retrained LLaMA-1B without BOS token

Figures 15 - 20 illustrates the self attention distribution for our LLaMA-1B that we trained from scratch. BOS token is *not* included in all of these visualizations.

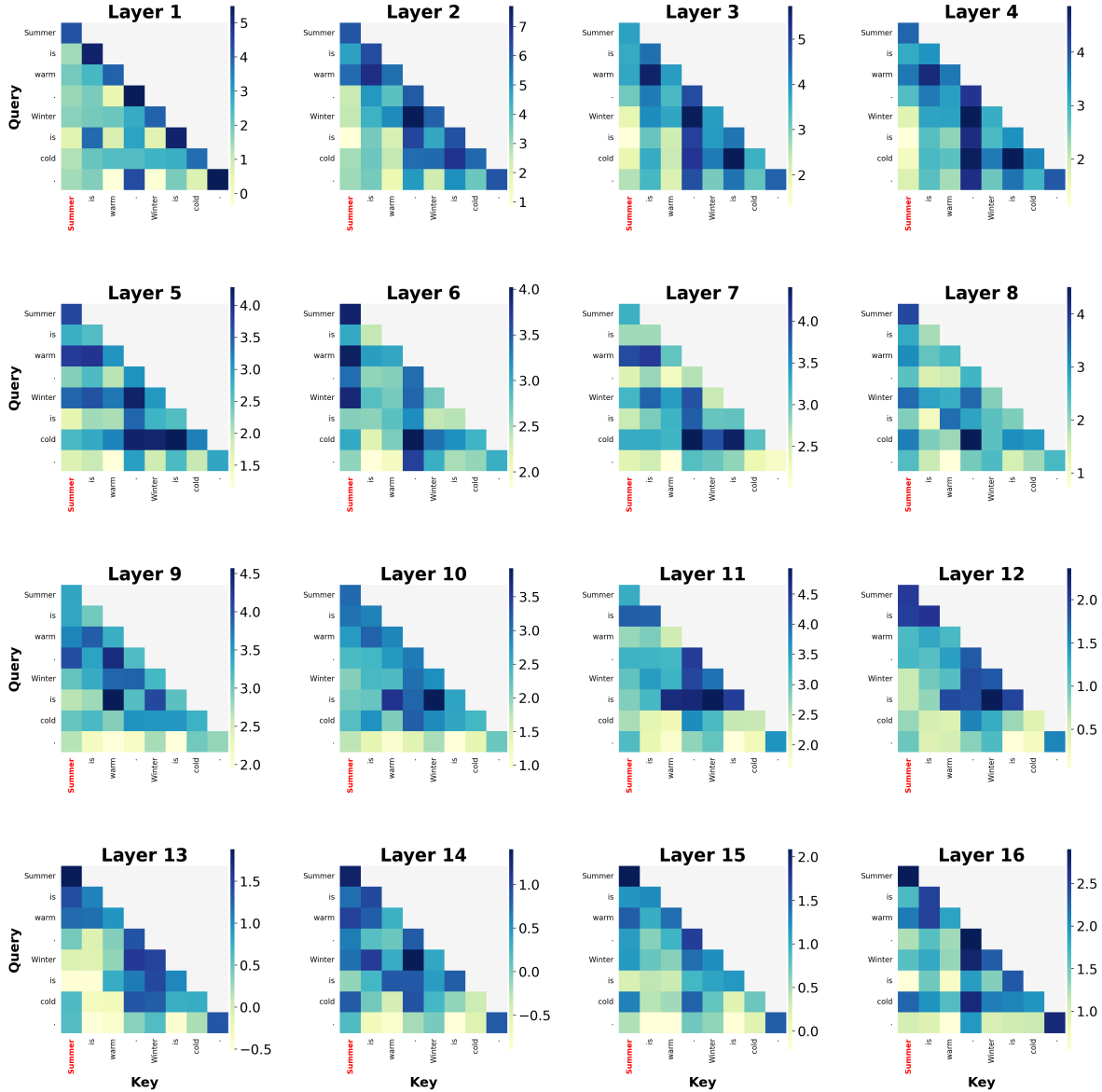

Figure 15: **Average attention logits over all heads for LLaMA-1B (baseline) without BOS token.**

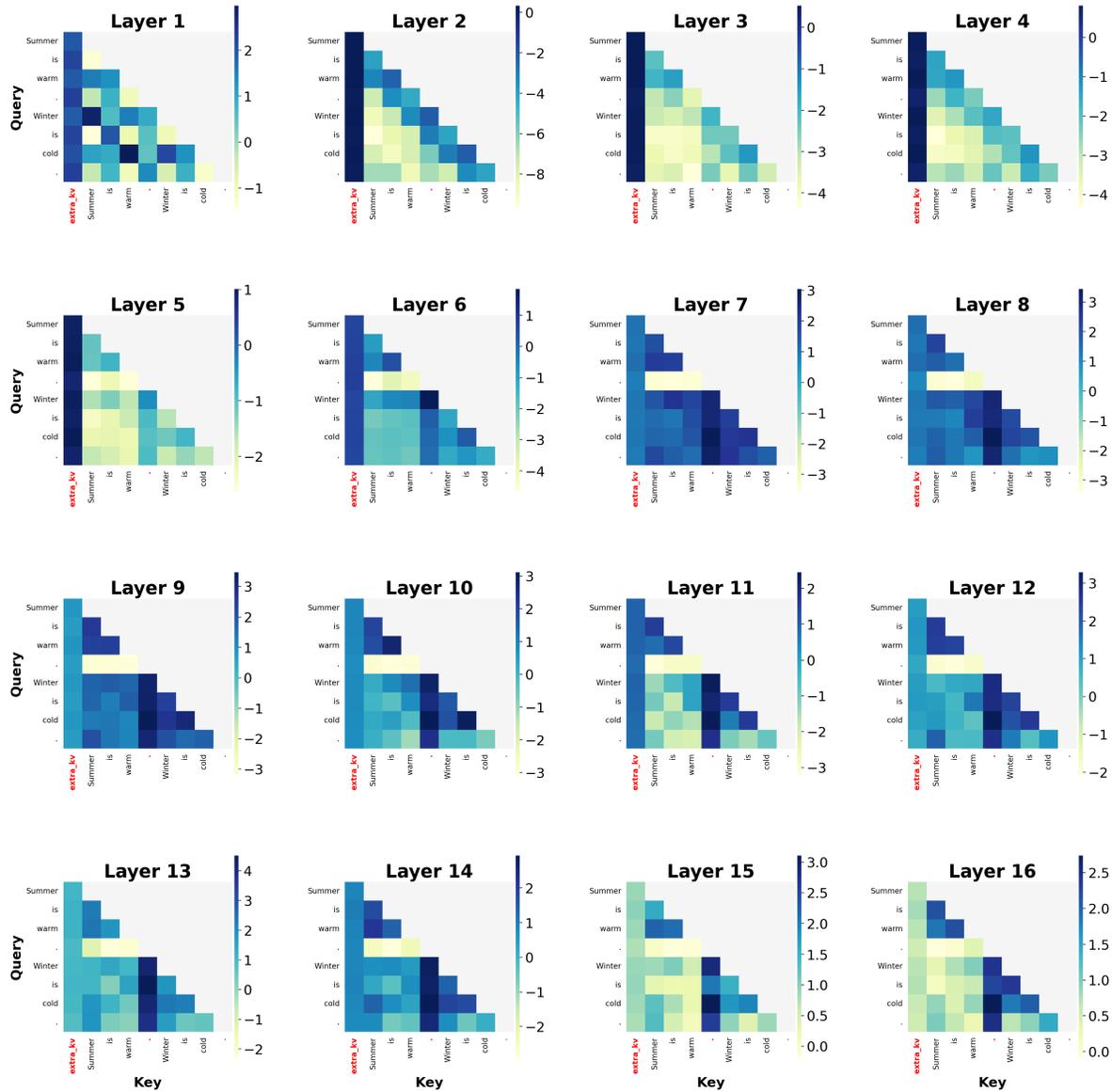

Figure 16: **Average attention logits over all heads for LLaMA-1B (with KV Bias) without BOS token.**

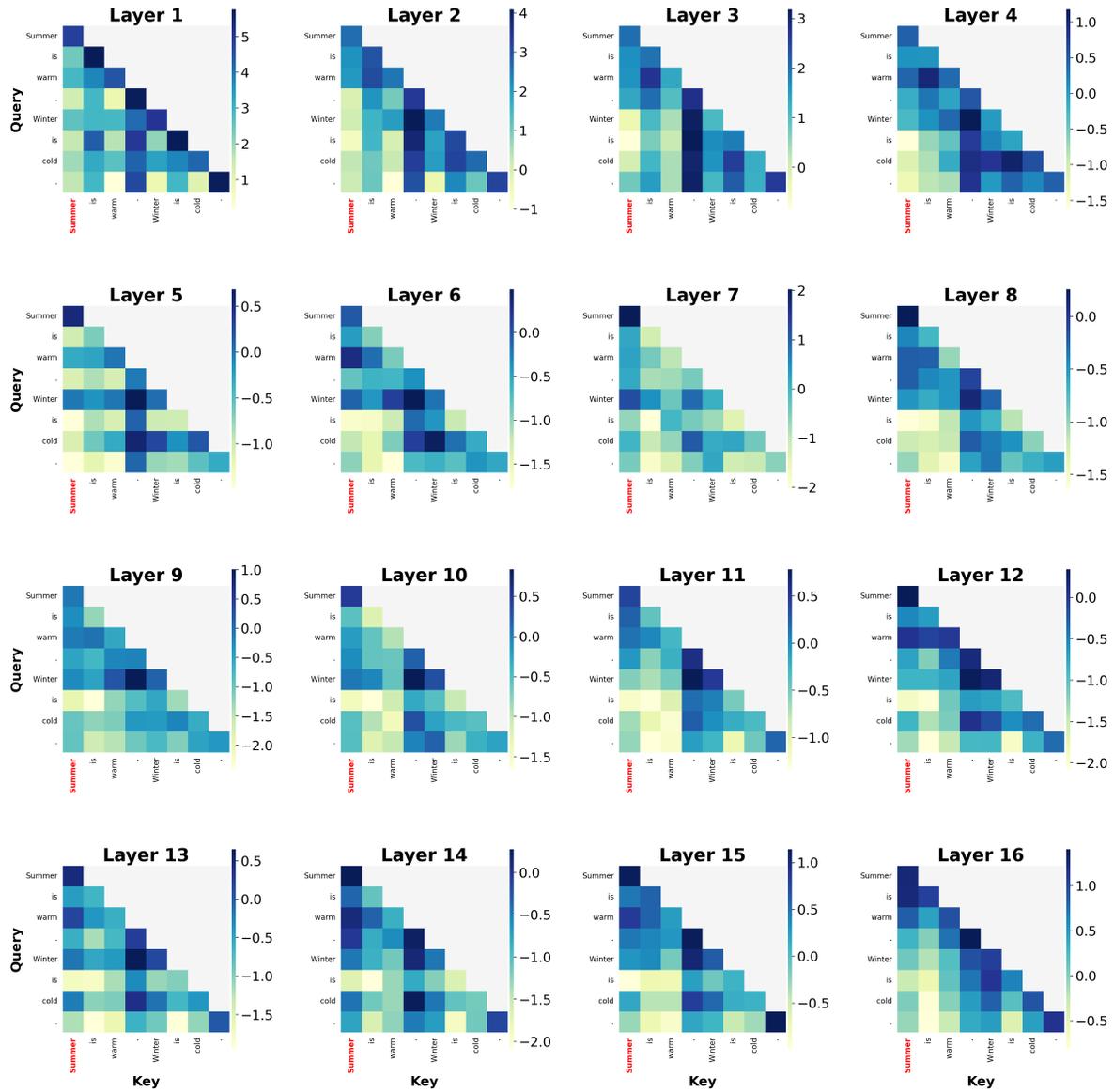

Figure 17: **Average attention logits over all heads for LLaMA-1B (with TVR) without BOS token.**

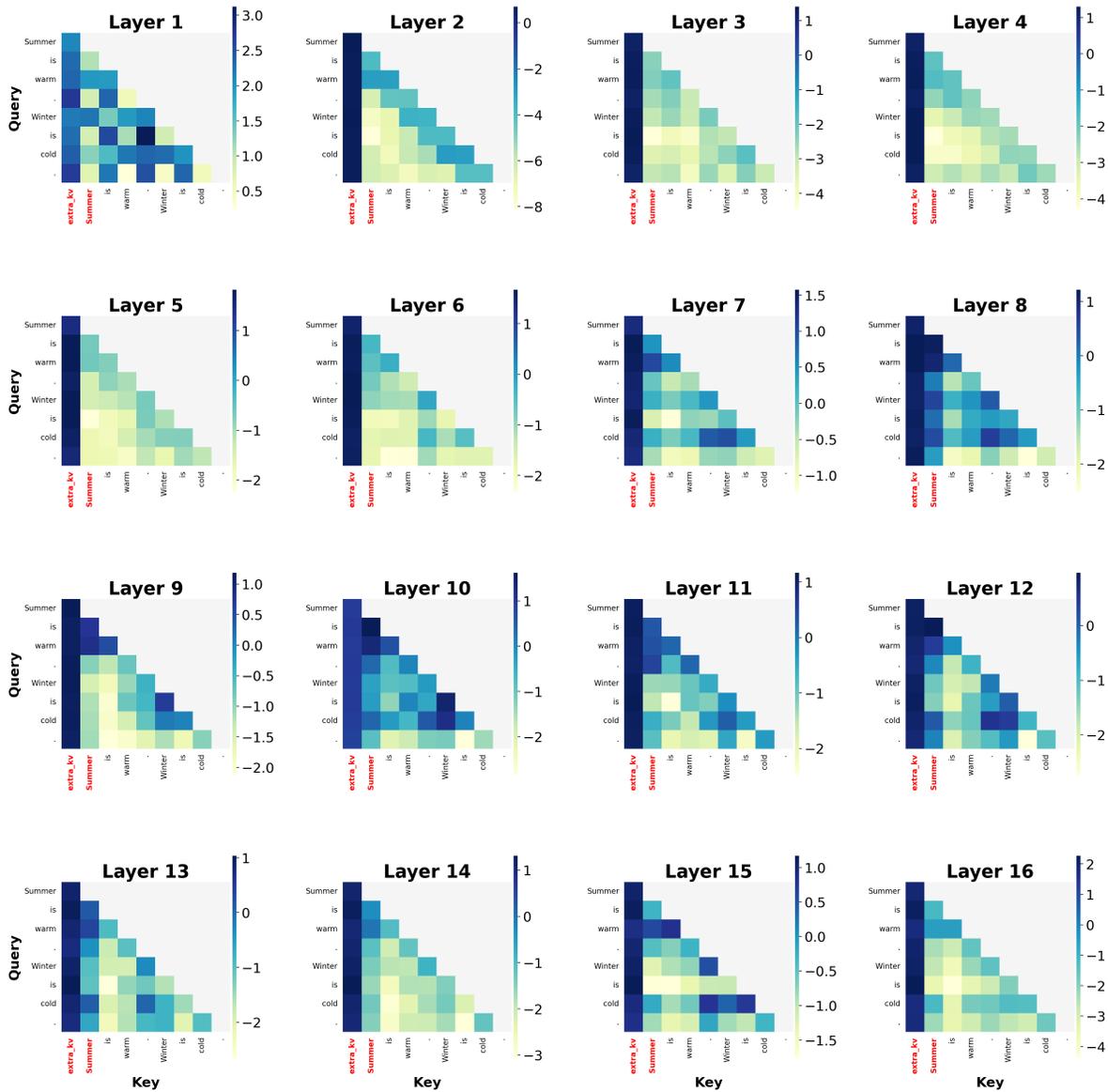

Figure 18: **Average attention logits over all heads for LLaMA-1B (with KV Bias + TVR) without BOS token.**

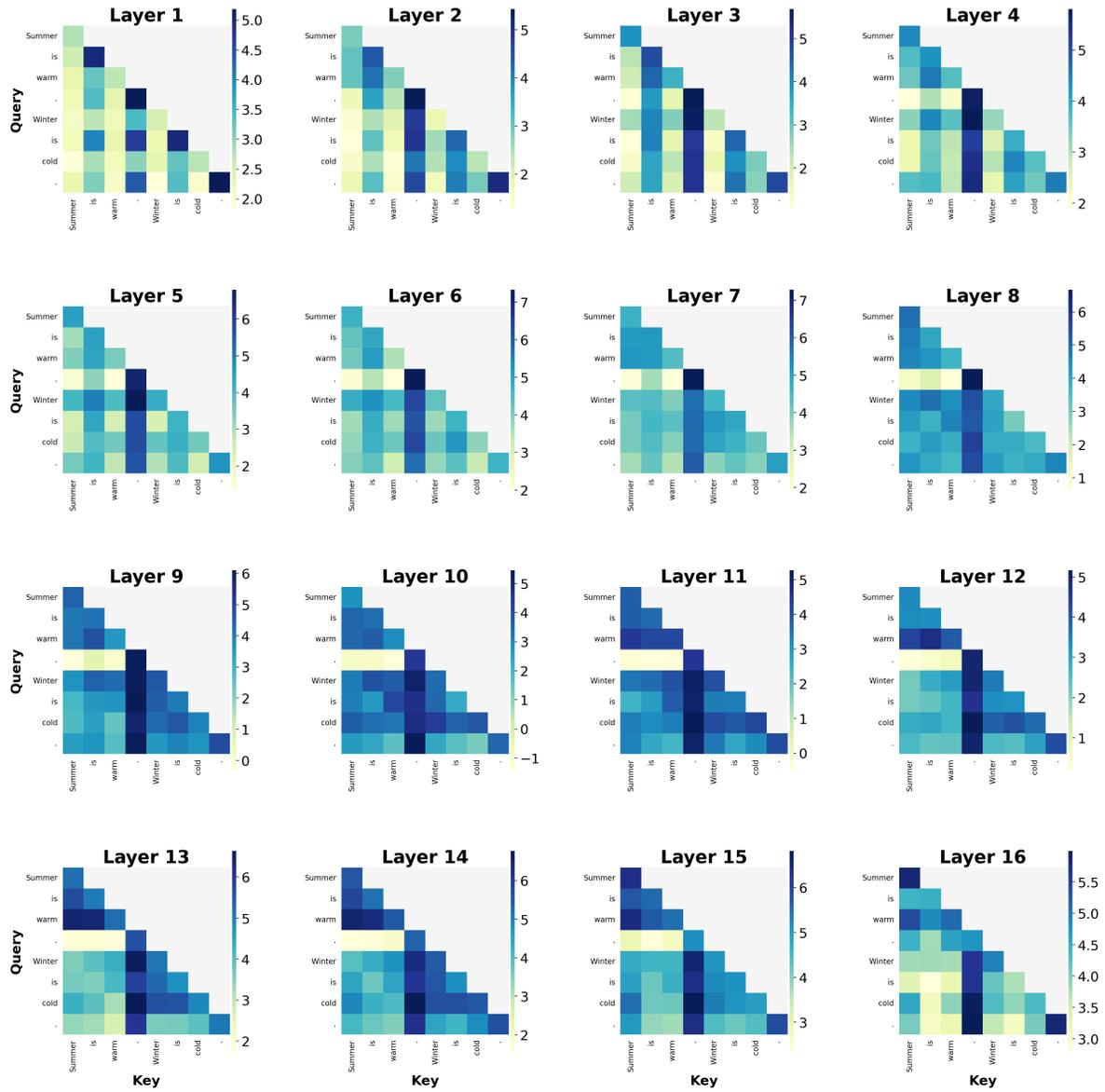

Figure 19: **Average attention logits over all heads for LLaMA-1B (with DyT) without BOS token.**

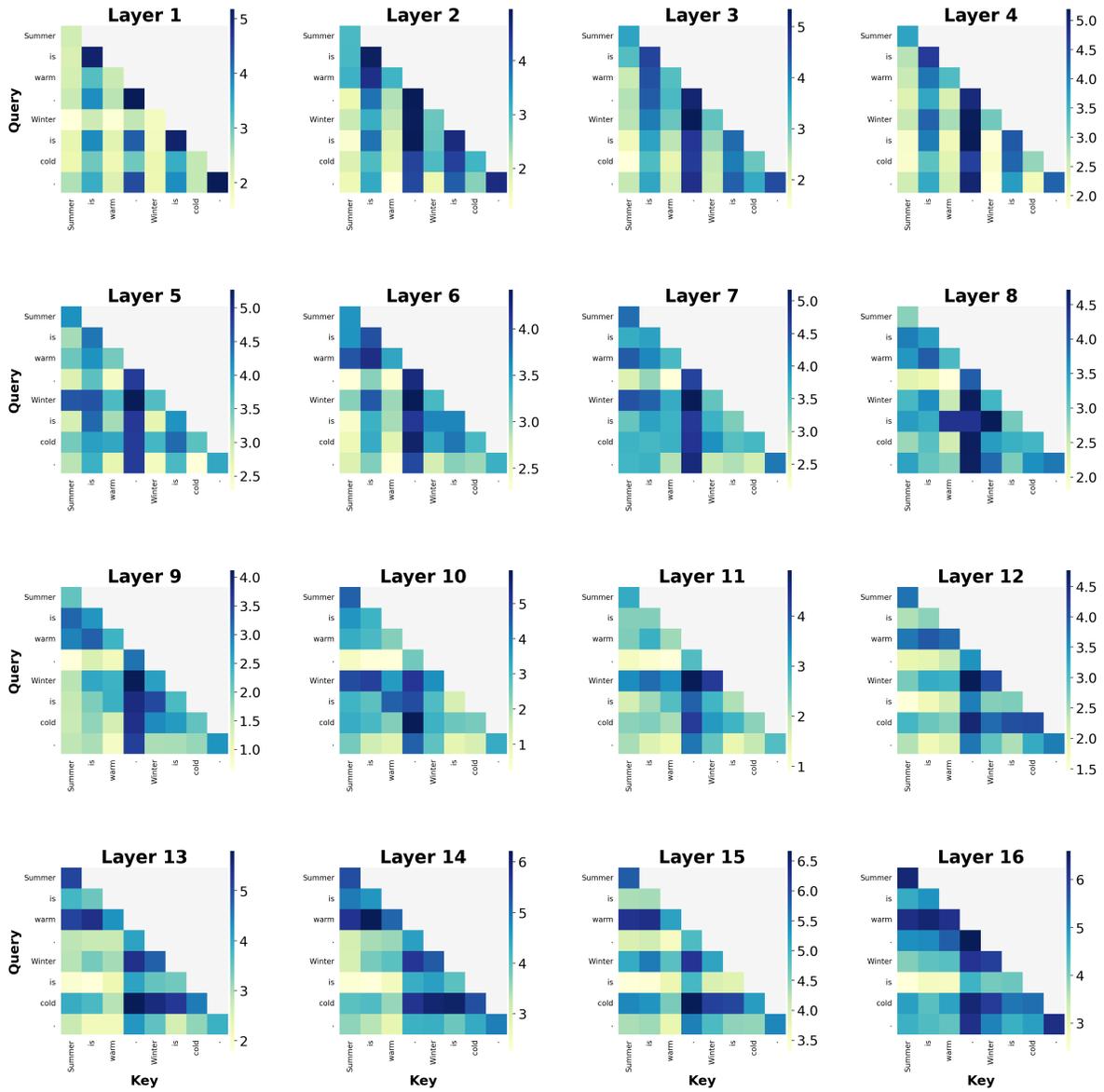

Figure 20: **Average attention logits over all heads for LLaMA-1B (with DyT + TVR) without BOS token.**

### 6.7.2 Self Attention Plots for Retrained LLaMA-1B with BOS token

Figures 21 - 26 illustrates the self attention distribution for our LLaMA-1B that we trained from scratch. BOS token is included in all of these visualizations.

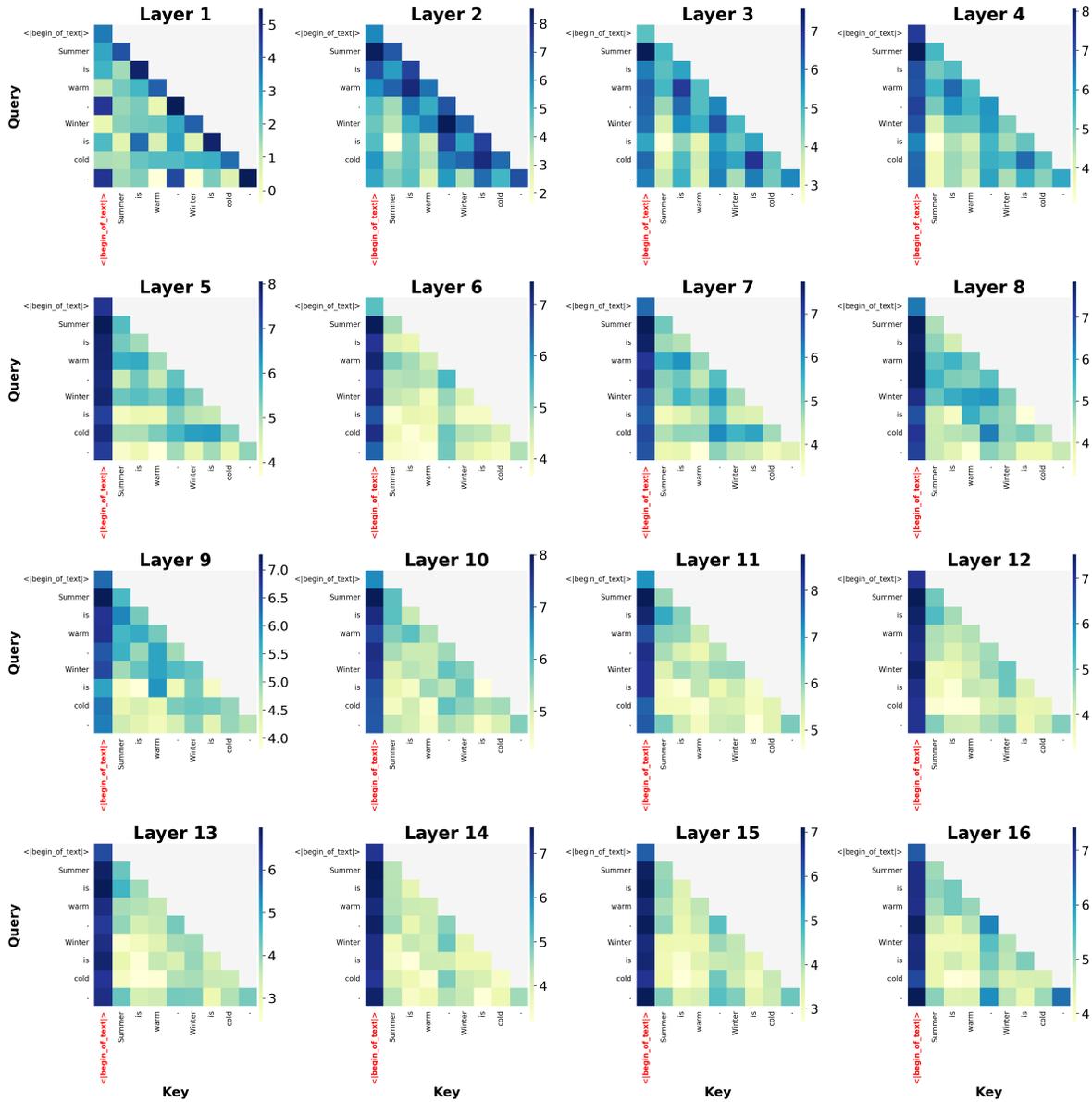

Figure 21: **Average attention logits over all heads for LLaMA-1B (baseline) with BOS token.**

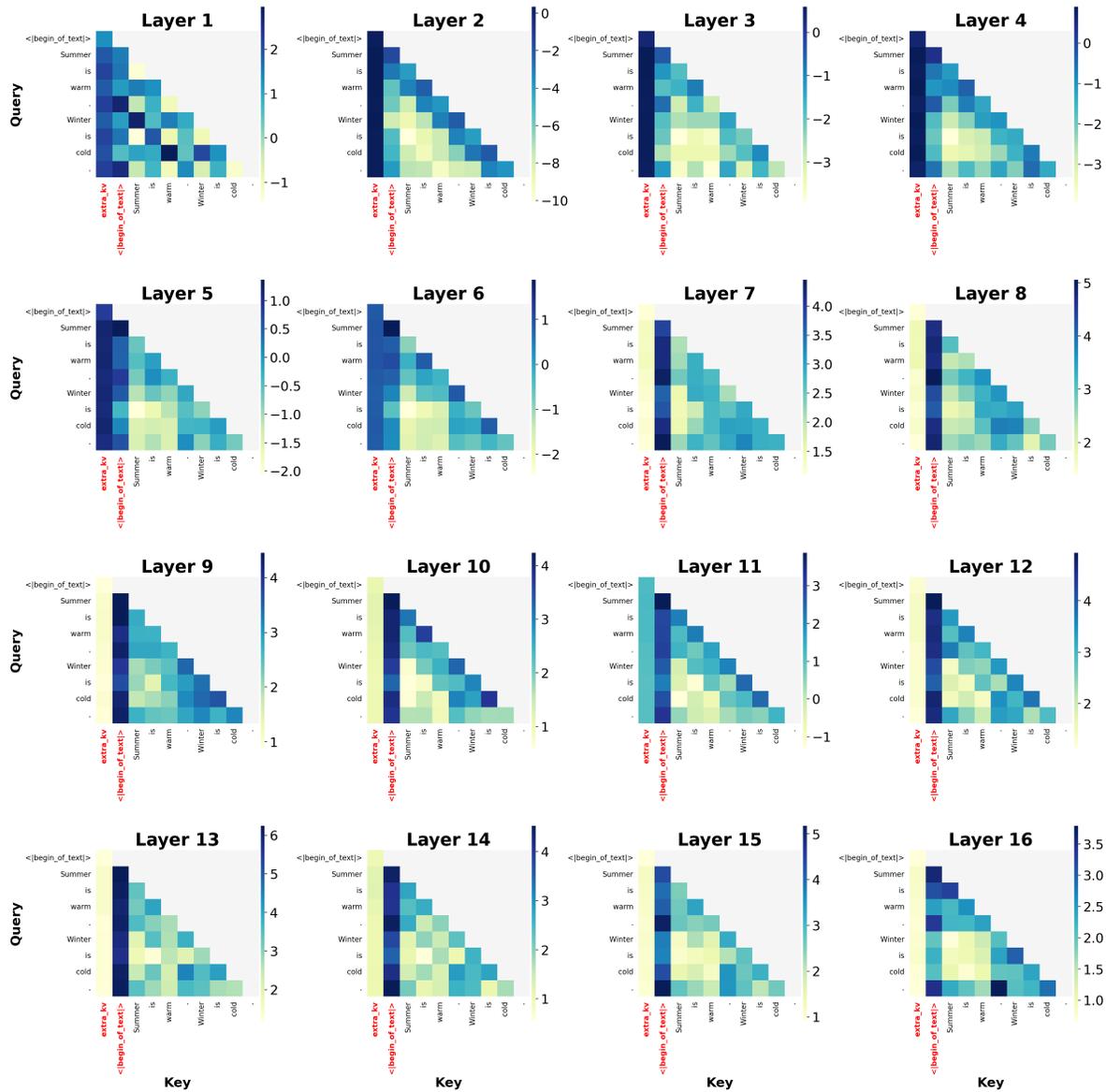

Figure 22: **Average attention logits over all heads for LLaMA-1B (with KV Bias) with BOS token.**

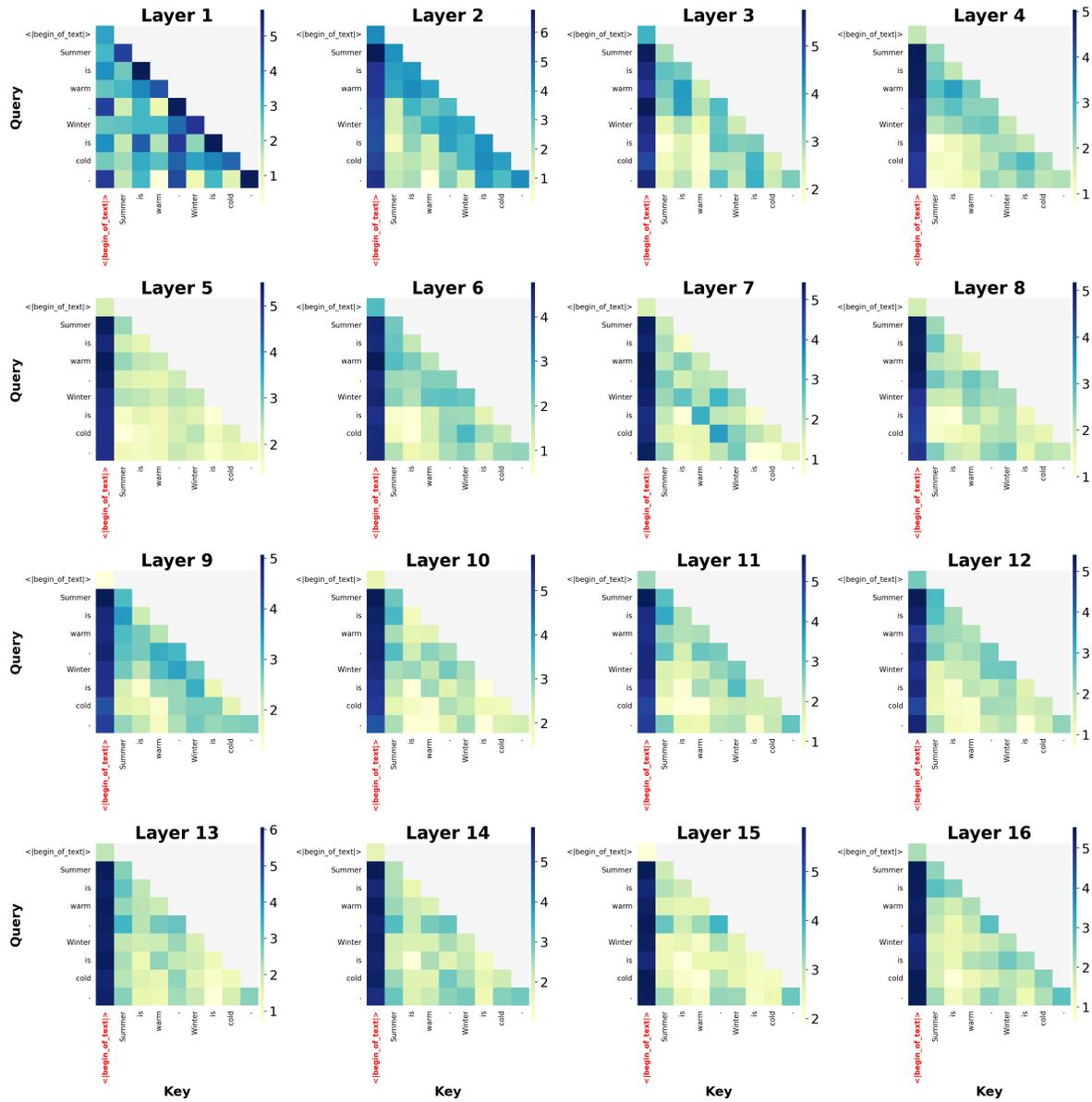

Figure 23: **Average attention logits over all heads for LLaMA-1B (with TVR) with BOS token.**

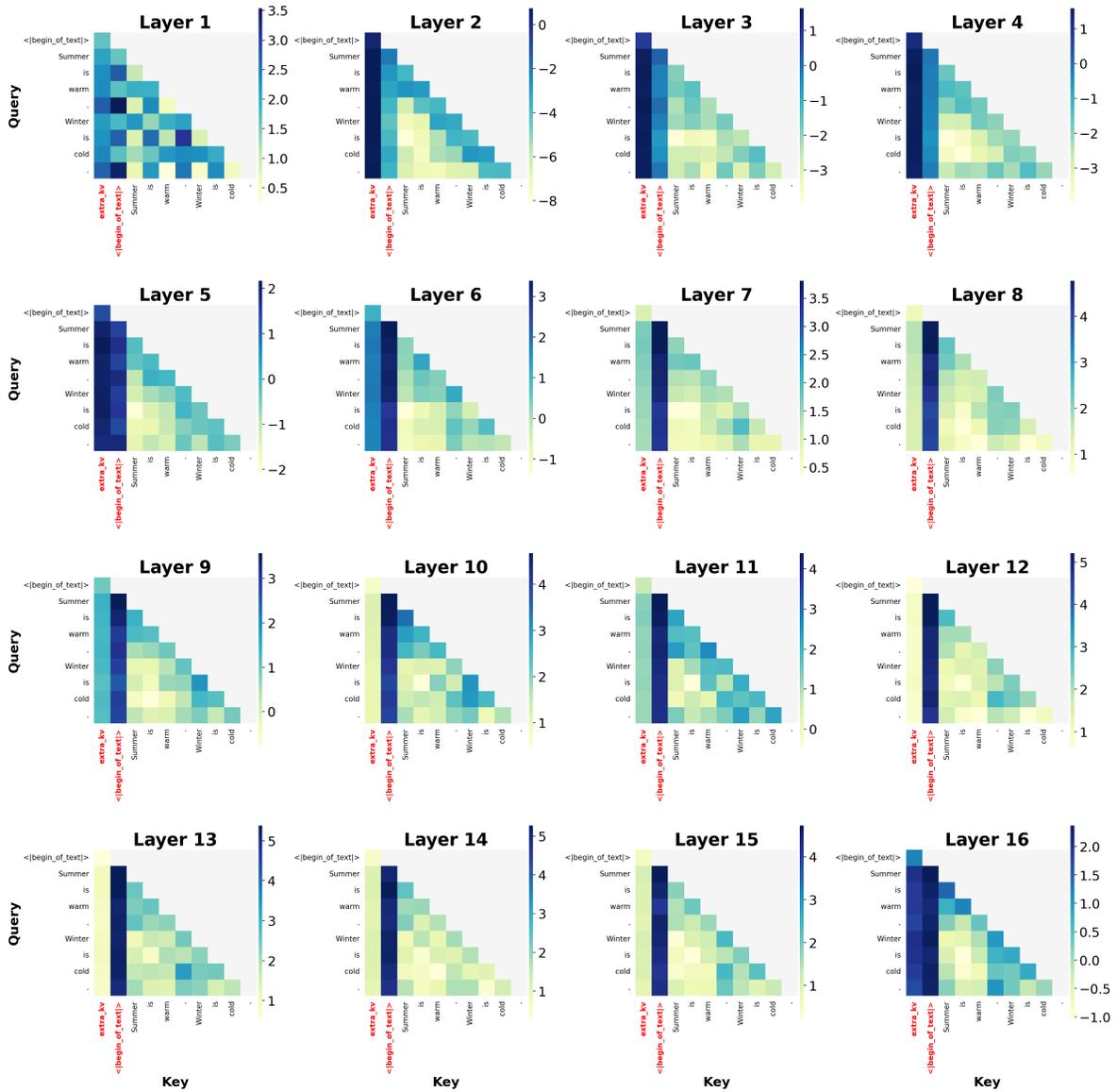

Figure 24: **Average attention logits over all heads for LLaMA-1B (with KV Bias + TVR) with BOS token.**

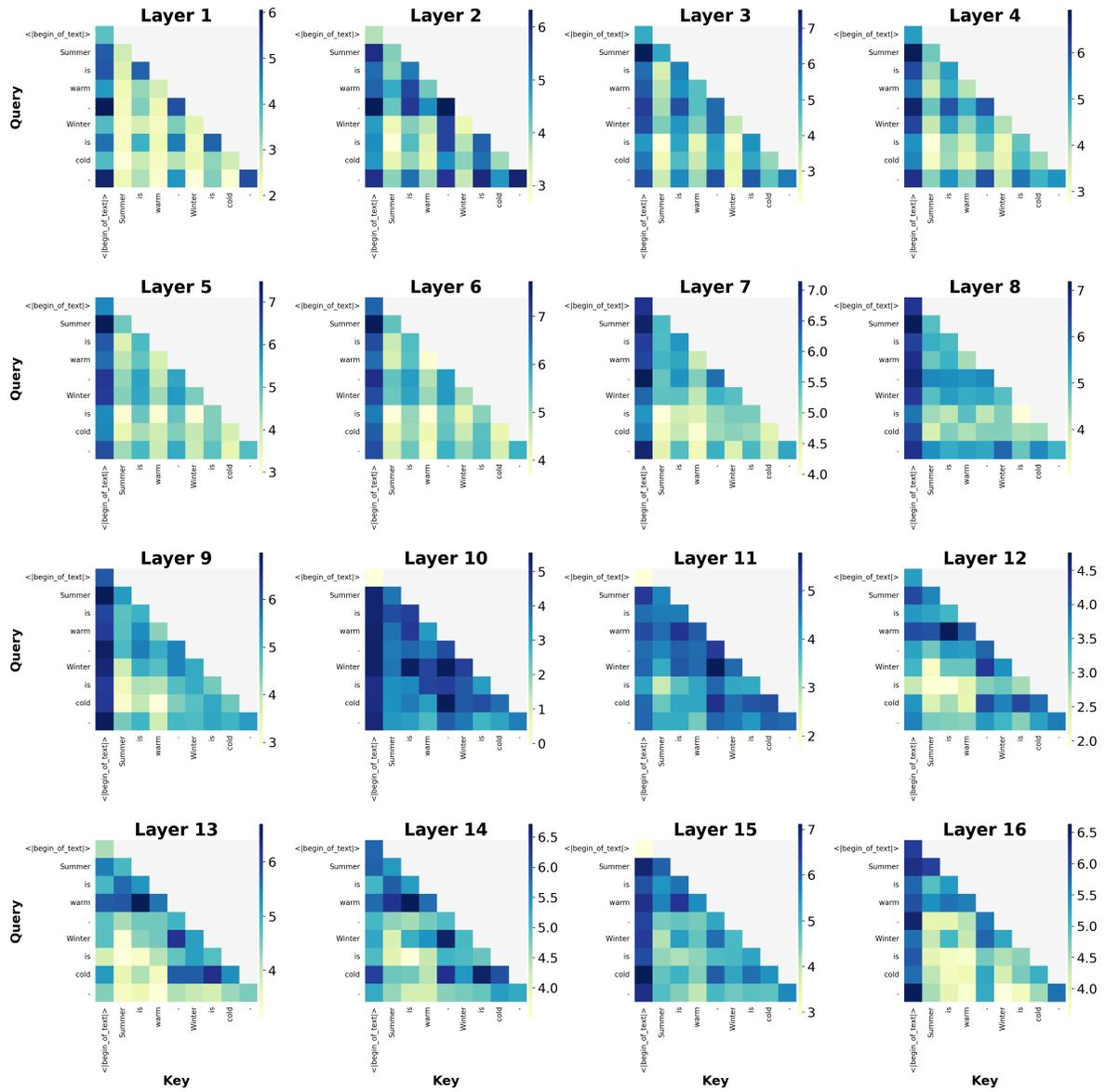

Figure 25: **Average attention logits over all heads for LLaMA-1B (with DyT) with BOS token.**

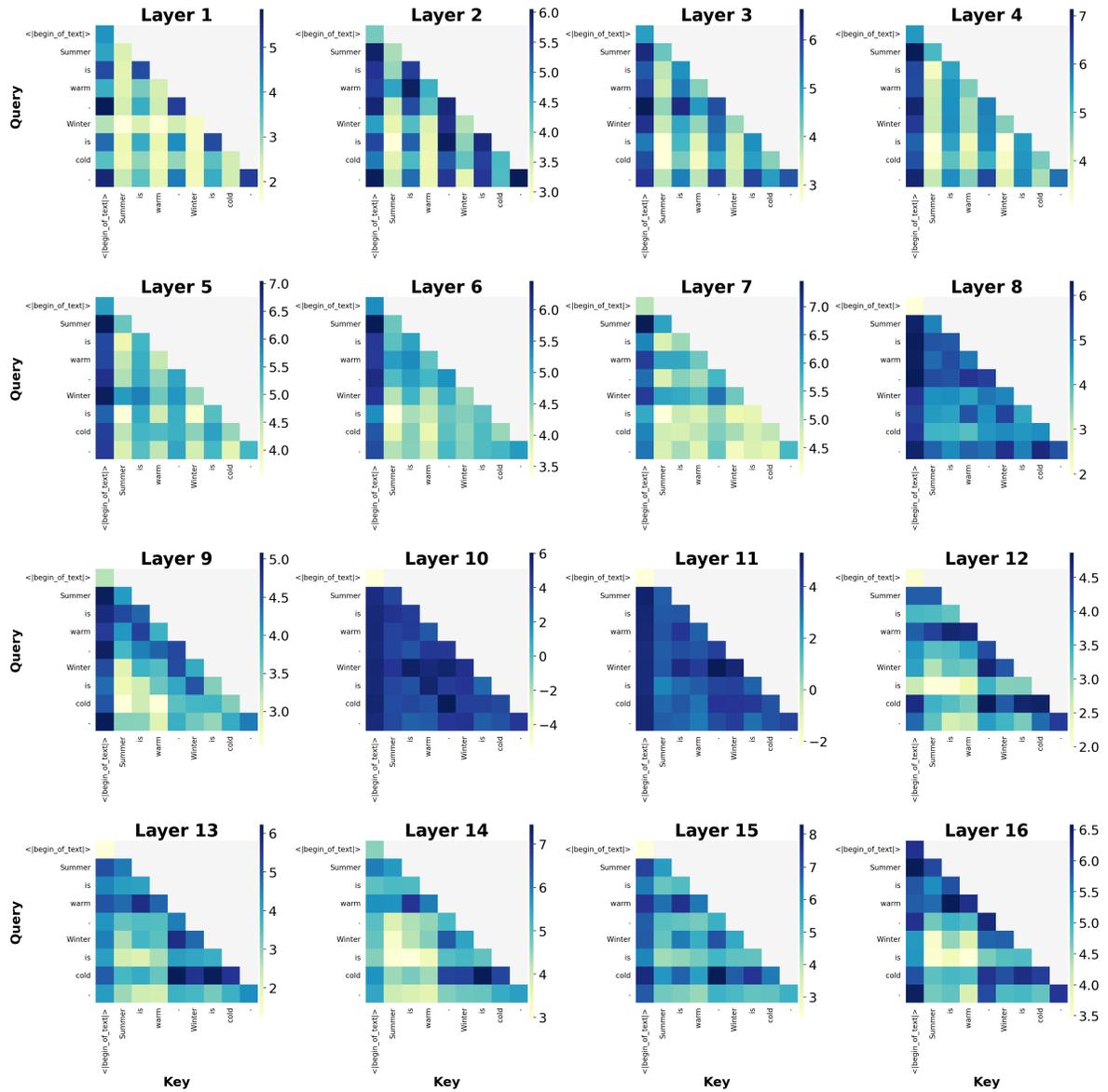

Figure 26: **Average attention logits over all heads for LLaMA-1B (with DyT + TVR) with BOS token.**

### 6.7.3 Top Activation Magnitude Plots for Retrained LLaMA-1B without BOS token

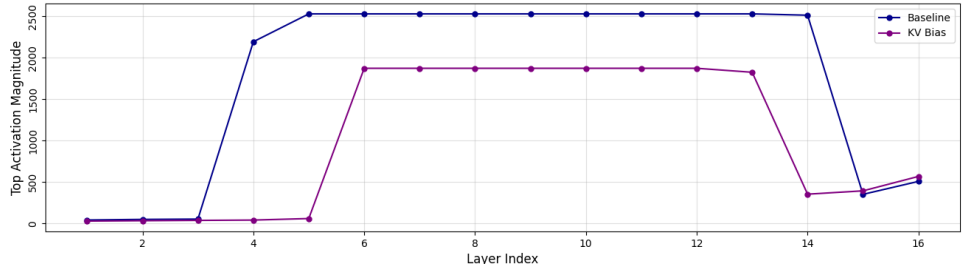

Figure 27: **Comparison of top activation magnitude across layers for LLaMA-1B between baseline and with KV Bias.** The input sentence is "Summer is warm. Winter is cold" without BOS token included.

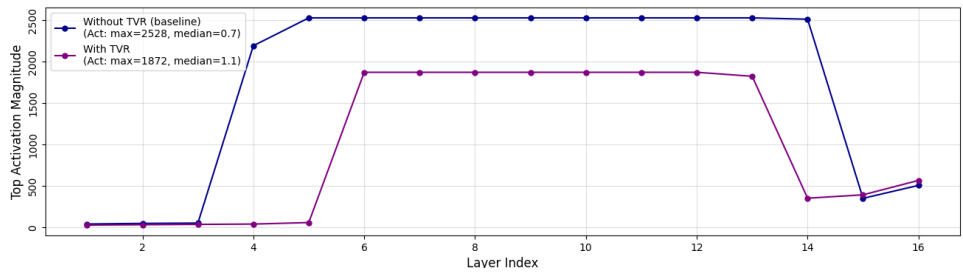

Figure 28: **Comparison of top activation magnitude across layers for LLaMA-1B without TVR (baseline) and with TVR.** The input sentence is "Summer is warm. Winter is cold" without BOS token included.

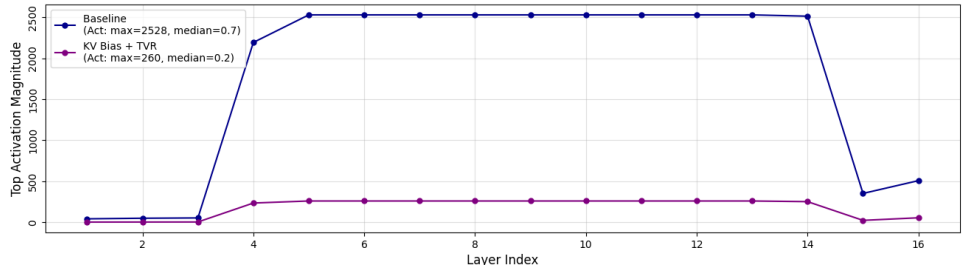

Figure 29: **Comparison of top activation magnitude across layers for LLaMA-1B between baseline and with KV Bias + TVR.** The input sentence is "Summer is warm. Winter is cold" without BOS token included.

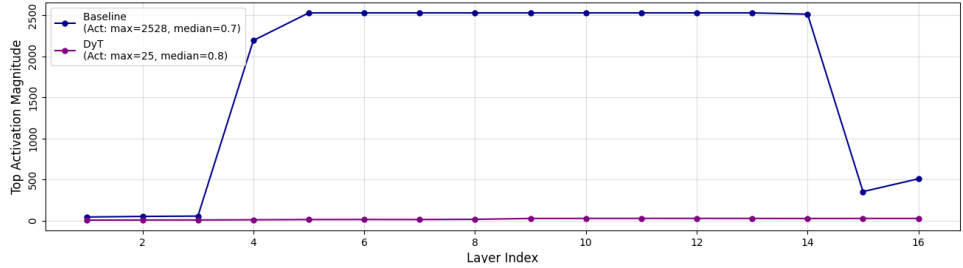

Figure 30: **Comparison of top activation magnitude across layers for LLaMA-1B between baseline and DyT**. The input sentence is "Summer is warm. Winter is cold" without BOS token included.

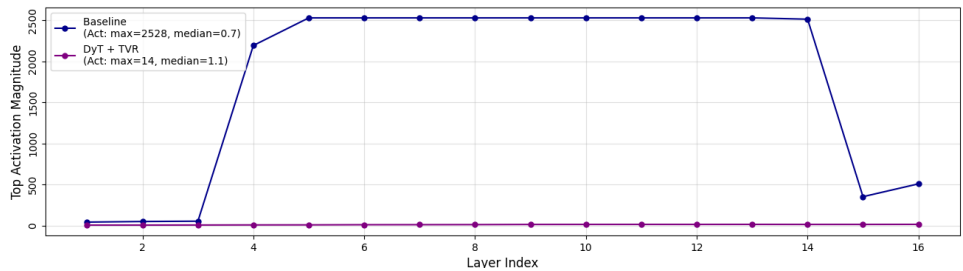

Figure 31: **Comparison of top activation magnitude across layers for LLaMA-1B between baseline and DyT + TVR**. The input sentence is "Summer is warm. Winter is cold" without BOS token included.

## 6.8 Breakdown of Downstream Tasks Performance for Various Mitigation Strategies

Table 7 contains the breakdown of downstream tasks performance for various mitigation strategies.

|  | HellaS | PIQA | SIQA | WinoG | TQA | ARC-E | ARC-C | **Mean** | $\Delta_{\textbf{base}}$ |
|---|---|---|---|---|---|---|---|---|---|
| Baseline | 52.5 | 71.2 | 40.1 | 53.6 | 39.2 | 61.8 | 33.9 | 50.3 | 0.0 |
| KV Bias | 50.3 | 71.5 | 41.3 | 53.6 | 35.5 | 62.6 | 32.3 | 49.6 | -0.7 |
| DyT | 50.7 | 71.2 | 40.0 | 51.3 | **39.8** | 63.0 | 33.0 | 49.9 | -0.4 |
| TVR | **55.1** | **73.4** | **42.7** | 57.0 | 38.9 | **64.3** | 36.1 | **52.5** | **2.2** |
| KV Bias + TVR | 54.8 | 72.5 | 42.0 | **57.9** | 37.9 | 62.4 | **36.3** | 52.0 | 1.7 |
| DyT + TVR | 50.9 | 70.9 | 41.7 | 51.7 | 40.1 | 61.7 | 34.8 | 50.3 | 0.0 |

Table 7: **Downstream tasks performance benchmark for different massive activation mitigation strategies on LLaMA-1B.**